\documentclass{ar-1col}

\usepackage{mathtools} 
\usepackage{soul}      
\usepackage{nameref}
\usepackage[comma]{natbib}
\usepackage{url}
\usepackage{subfig}
\usepackage{amssymb,amsmath}
\usepackage{bm}
\usepackage{bbm} 
\usepackage{tikz}
\usepackage{graphicx}
\usepackage{xr}
\usetikzlibrary{decorations.pathreplacing}

\jname{Annu.~Rev.~Stat.~Appl.}
\jvol{(Vol)}
\jyear{2025}
\doi{10.1146/(DOI)}

\newcommand{\proglang}[1]{\texttt{#1}}
\newcommand{\pkg}[1]{{\fontseries{b}\selectfont #1}}

\definecolor{ARSIAred}{HTML}{8f2a29}

\newcommand{\KL}[2]{\textrm{KL}\{#1~\|~#2\}}
\newcommand{\KLb}[2]{\textrm{KL}[#1~\|~#2]}
\newcommand{\zerob} {{\bf 0}}

\newcommand{\thetab} {{\boldsymbol{\theta}}}
\newcommand{\mub} {{\boldsymbol{\mu}}}

\newcommand{\kappab} {{\boldsymbol{\kappa}}}

\newcommand{\nub} {{\boldsymbol{\nub}}}
\newcommand{\gammab} {{\boldsymbol{\gamma}}}
\newcommand{\deltab} {{\boldsymbol{\delta}}}

\newcommand{\etab} {{\boldsymbol{\eta}}}

\newcommand{\thetahat} {{\hat{\theta}}}
\newcommand{\thetabhat} {{\widehat{\thetab}}}

\newcommand{\intd} {\textrm{d}}

\newcommand{\zetab} {\boldsymbol{\zeta}}

\newcommand{\Sigmamat} {{\boldsymbol \Sigma}}

\newcommand{\Lmat} {\textbf{L}}

\newcommand{\Smat} {\textbf{S}}
\newcommand{\Zmat} {\textbf{Z}}

\newcommand{\Xvec} {\mathbf{X}}
\newcommand{\Svec} {\mathbf{S}}
\newcommand{\Imat} {\textbf{I}}

\newcommand{\svec} {\textbf{s}}

\newcommand{\muvec} {\boldsymbol{\mu}}

\newcommand{\taub} {\boldsymbol {\tau}}
\newcommand{\omegab} {\boldsymbol {\omega}}

\renewcommand{\zerob}{\mathbf{0}}

\newcommand{\Yvec}{\mathbf{Y}}

\newcommand{\Zvec}{\mathbf{Z}}

\newcommand{\E}{\mathbb{E}}

\newcommand{\vech}{\mathrm{vech}}

\DeclareMathOperator*{\argmin}{arg\,min}
\DeclareMathOperator*{\argmax}{arg\,max}

\let\originalleft\left
\let\originalright\right
\renewcommand{\left}{\mathopen{}\mathclose\bgroup\originalleft}
\renewcommand{\right}{\aftergroup\egroup\originalright}

\begin{document}

\markboth{Zammit-Mangion et al.}{Neural Methods for Amortized Inference}

\title{Neural Methods for Amortized Inference}

\author{Andrew Zammit-Mangion$^1$, \\ Matthew Sainsbury-Dale$^{1,2}$, and \\ Rapha{\"e}l Huser$^2$
\affil{$^1$School of Mathematics and Applied Statistics, University of Wollongong, Wollongong, New South Wales, Australia; email: azm@uow.edu.au}
\affil{$^2$Statistics Program, Computer, Electrical and Mathematical Sciences and Engineering Division, King Abdullah University of Science and Technology (KAUST), Thuwal, Saudi Arabia}}
\begin{abstract}
Simulation-based methods for statistical inference have evolved dramatically over the past 50 years, keeping pace with technological advancements. The field is undergoing a new revolution as it embraces the representational capacity of neural networks, optimization libraries and graphics processing units for learning complex mappings between data and inferential targets. The resulting tools are amortized, in the sense that, after an initial setup cost, they allow rapid inference through fast feed-forward operations. In this article we review recent progress in the context of point estimation, approximate Bayesian inference, summary-statistic construction, and likelihood approximation. We also cover software, and include a simple illustration to showcase the wide array of tools available for amortized inference and the benefits they offer over Markov chain Monte Carlo methods. The article concludes with an overview of relevant topics and an outlook on future research directions.
\end{abstract}

\maketitle

\section{INTRODUCTION}\label{sec:intro}

Statistical inference, the process of drawing conclusions on an underlying population from observations,  is a cornerstone of evidence-based decision making and scientific discovery. It often relies on a statistical model with unknown or uncertain parameters. Parameter inference consists in estimating and quantifying uncertainty over these model parameters. When the underlying model admits a likelihood function that is analytically and computationally tractable, likelihood-based methods are available that are well-suited for inference. However, in many cases the likelihood function is either unavailable or computationally prohibitive to evaluate, but it is feasible to simulate from the model; this is the case, for example, with many geophysical models \citep[e.g.,][]{Siahkool_2023}, epidemiological models \citep{Fasiolo_2016}, cognitive neuroscience models \citep{Fengler_2021}, agent-based models \citep{Dyer_2024}, 
and some classes of statistical models like Markov random fields \citep[e.g.,][]{Besag_1986} or models for spatial extremes \citep[e.g.,][]{Davison_2012,Huser_2022}. In such cases, one often resorts to simulation-based techniques to bypass evaluation of the likelihood function. Simulation-based inference requires substantial computing capability, and thus only became viable in the second half of the twentieth century  (for early examples, see \citeauthor{Hoel_1971} \citeyear{Hoel_1971} and \citeauthor{Ross_1971} \citeyear{Ross_1971}). The field has evolved much since then, due to the dramatic increase in affordable computing power, and the increased ability to generate, store, and process large datasets.

\begin{marginnote}
    \entry{Simulation-based inference}{Any procedure that makes statistical inference using simulations from a generative model.}
 \end{marginnote}

Several review articles have been published in recent years on simulation-based inference. \citet{Cranmer_2020} give a succinct summary of several methods ranging from approximate Bayesian computation \citep[ABC;][]{Sisson_2018_ABC_handbook} to density estimation \citep{Diggle_1984} and more recent machine learning approaches, including some of the neural network-based approaches we review here. \citet{Blum_2013} and \citet{Grazian_2020} focus on ABC methods, while \citet{Drovandi_2018} and \citet{Drovandi_2022} additionally review indirect inference \citep{Gourieroux_1993_indirect_inference}, and Bayesian synthetic likelihood \citep{Wood_2010} approaches to simulation-based inference.

  This article differs from existing reviews on simulation-based inference in two ways.
First, it only considers methods that incorporate neural  networks, 
which have become state-of-the-art in high-dimensional modeling due to their representational capacity \citep{Hornik_1989} and due to the increased availability of the software and hardware required to train them.
Second, it focuses on neural simulation-based methods that are amortized, that is, that leverage a feed-forward relationship between the data and the inferential target to allow fast inference. The Oxford English Dictionary defines amortization as ``the action or practice of gradually writing off the initial cost of an asset''. In the context of neural amortized inference, the initial cost involves training a neural network to learn a complex mapping used for inference (a point estimator, an approximate posterior distribution, an approximate likelihood function, etc.). Once the neural network is trained, parameter inferences can then be made several orders of magnitude faster than classical methods such as Markov chain Monte Carlo (MCMC) sampling. Hence, the initial cost of training the network is ``amortized over time'', as the trained network is used over and over again for inferences with new data (see the sidebar titled The Power of Amortization). Amortized inference is also associated with the way humans operate, reusing demanding inferences made in the past to make quick decisions \citep{Gershman_2014}.\begin{marginnote}
    \entry{Neural network}{A flexible, highly parameterized nonlinear mapping, constructed using a composition of simple functions connected together in a network.}
    \entry{Training}{The process of estimating (or ``learning'') parameters in a neural network, typically by minimizing an expected loss.}
  \end{marginnote}

\begin{textbox}[t!]\section{THE POWER OF AMORTIZATION} Neural networks are ideally suited for situations in which a complex task is repeated, and where the problem justifies a potentially substantial initial outlay. In such settings, one expends a cost to initially train the network with the intention of reaping dividends over time with its repeated use, a strategy known as amortization. The power of amortization is perhaps most clearly exhibited in large language models. Training the BigScience Large Open-science Open-access Multilingual Language Model (BLOOM) required over a million hours of computing on several hundred state-of-the-art graphics processing units (GPUs), each costing many thousands of dollars, for nearly four months \citep{Luccioni_2023}. The total estimated consumed energy for training was 433 MWh, which is roughly equivalent to the yearly consumption of 72 Australian households. However, once trained, BLOOM could repeatedly produce polished text outputs (inferences) in response to inputs (data) on a single GPU with a total consumed energy far smaller than that needed to boil water for a cup of tea. \end{textbox}

Not all simulation-based inference methods establish a fast-to-evaluate feed-forward relationship between the data and the inferential target. These non-amortized methods require re-optimization or re-simulation every time new data are observed. This makes them unsuitable for situations where the same inferential task must be repeated many times on different datasets, as is often the case in operational settings. For example, NASA's Orbiting-Carbon-Observatory-2 satellite takes approximately 1,000,000 measurements of light spectra per day. For each spectrum, an estimate of carbon dioxide mole fraction is obtained by computing a  Bayesian posterior mode using a forward physics model \citep{Cressie_2018}, requiring an exorbitant amount of computing power when done at scale. Neural networks offer a way to make Bayesian inference from the spectra quickly and accurately at a much lower computational cost \citep{David_2021}.

\subsection{Article outline}
This review provides an introductory high-level roadmap to amortized neural inferential methods. It organizes and categorizes these methodologies to help navigate this relatively new and dynamic field of research. In Section~\ref{sec:Brown} we describe amortization from a decision-theoretic perspective. In Section~\ref{sec:neural_posterior_inference} we discuss (Bayesian) point estimation and methods that yield approximate posterior distributions.  
In Section~\ref{sec:summarystatistics} we show how neural networks can be used to construct summary statistics, which often feature in amortized statistical inference methods. In Section~\ref{sec:likelihood_and_ratio}, we review neural methods for amortized likelihood-function and likelihood-to-evidence-ratio approximation. In Section~\ref{sec:software}, we discuss software for making amortized parameter inference, and we demonstrate its use on a simple model where asymptotically exact inference is possible. The Supplemental Appendix contains a second software example, further reading on topics related to the main subject of the review, and additional figures and tables.

\subsection{Notation convention}
We use boldface to denote vectors and matrices, $\deltab$ and $\deltab(\cdot)$ to denote a generic decision and decision rule, respectively, and $\Xvec \in \mathcal{X}$ to denote a generic random vector that takes values in  $\mathcal{X}$. We denote data from a sample space $\mathcal{Z}$  as $\Zvec \equiv (Z_1,\dots,Z_n)' \in \mathcal{Z} \subseteq \mathbb{R}^n$, and a model parameter vector of interest as $\thetab \equiv (\theta_1,\dots,\theta_d)' \in \Theta$, where $\Theta \subseteq \mathbb{R}^d$ is the parameter space.  Unless specified otherwise, we assume that all measures admit densities with respect to the Lebesgue measure. To simplify notation, 
we let the argument of a density function $p(\cdot)$ (which, from hereafter, we simply refer to as a distribution) indicate both the random variable the distribution is associated with and its input argument. 
In particular, $p(\thetab)$ is the prior distribution of the parameters, $p(\Zvec)$ is the marginal likelihood (also known as the model evidence), and ${p(\thetab\mid \Zvec)}={p(\Zvec\mid \thetab)p(\thetab)/p(\Zvec)}$ is the posterior distribution of $\thetab \in \Theta$ given $\Zvec \in \mathcal{Z}$. We refer to $p(\Zvec\mid\thetab)$ as the likelihood function, whether it is expressing a distribution of $\Zvec$ for some fixed $\thetab$ or a function of $\thetab$ for some fixed $\Zvec$, with its definition taken from the context in which it is used. We denote an approximate function (e.g., approximate likelihood function or posterior distribution) generically as $q(\cdot)$; we use $\kappab$ to parameterize the approximation when $q(\cdot)$ refers to an approximate posterior distribution, and use $\omegab$ when $q(\cdot)$ refers to an approximate likelihood function. 
We often express these parameters as functions; for example, $\kappab_{\gammab}(\Zvec)$ is a function  parameterized by $\gammab$ that takes data $\Zvec$ as input and that outputs parameters characterizing the approximate posterior distribution of $\thetab$.

\section{AMORTIZATION: A DECISION-THEORETIC PERSPECTIVE}\label{sec:Brown}

Consider a random vector $\Xvec \in \mathcal{X}$. A decision rule $\deltab(\cdot)$ on $\mathcal{X}$ is a function that returns a decision for any given $\Xvec \in \mathcal{X}$. In statistical inference, the most well-known case is where $\Xvec$ is data and the decision rule is a point estimator returning a point estimate from data \citep{Casella_2001}. Decision rules are typically designed to optimize an objective function, such as a minimum-risk objective function. Optimal decision rules can rarely be expressed in closed form, and often require a computationally expensive optimization procedure to evaluate. \emph{Amortized} optimal decision rules are closed-form expressions that may require considerable modeling and computing effort to construct, but that subsequently yield approximately optimal decisions for any $\Xvec$ with little computing effort. They are central to amortized methods for quickly deriving point estimates, approximate posterior distributions, summary statistics, or approximate likelihood functions.

\begin{marginnote}
  \entry{Point estimator}{Any function of the data $\Zvec$ that returns an estimate of an unknown model parameter $\thetab$.}
\end{marginnote}

Finding an optimal decision often involves solving $\deltab^* = \inf_\deltab g(\Xvec, \deltab)$, where $g(\cdot,\cdot)$ is an objective function to minimize. For example, when $\deltab$ is a point estimate and $\Xvec$ is data, a popular choice for $g(\cdot,\cdot)$ is the posterior expected loss (Section~\ref{sec:neural_point_estimation}). Repeating this optimization procedure many times for different $\Xvec$ could be time-consuming or computationally prohibitive. \citet{Brown_1973} show that under mild conditions, including $g(\cdot,\cdot)$ being bounded, there exists a decision rule $\deltab^*(\cdot)$ that satisfies
\begin{equation}\label{eq:Brown}
g\{\Xvec, \deltab^*(\Xvec)\} = \inf_\deltab g(\Xvec,\deltab),\quad \mbox{for all}\;\;\Xvec \in \mathcal{X}.
\end{equation}
Equation~\ref{eq:Brown} provides the motivation for amortized methods: Once an accurate closed-form approximation to the decision rule $\deltab^*(\cdot)$ is found, computationally-intensive optimization routines are no longer needed. The computational burden is shifted from evaluation of $\deltab^*(\Xvec)$ for each $\Xvec$ to that of constructing an optimal mapping $\deltab^*(\cdot)$.

\begin{figure}[t!]
\centering
\captionsetup{width=\linewidth}
\includegraphics[width=\linewidth]{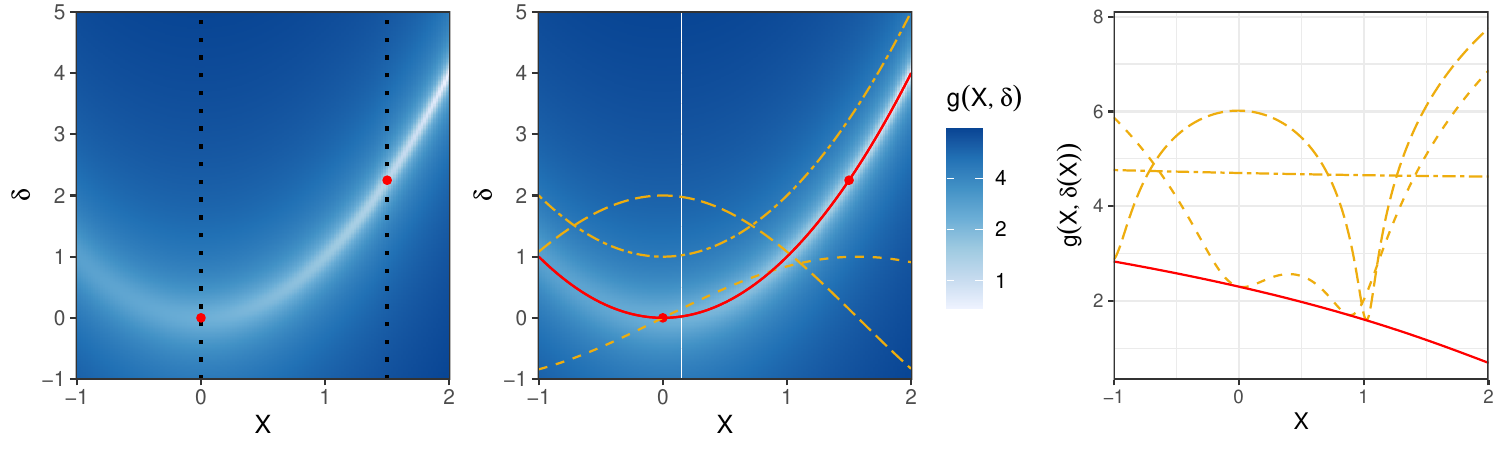}

\caption{(Left) A hypothetical nonnegative function  $g(X, \delta)$. To find the optimal decision $\delta^*$ for a given $X$ (shown as red dots for $X =0$ and $X = 1.5$), $g(X, {\delta})$ must be minimized along a slice at $X$ (black dotted lines).
 (Centre) A function $\delta^*(\cdot)$ (red solid line) that minimizes $g(\cdot, \cdot)$ for any $X \in \mathcal{X}$, whose existence is proved by \citet{Brown_1973}, and alternative decision rules (orange dashed lines). If known, or well-approximated, in closed form, $\delta^*(\cdot)$ can be used to quickly make optimal decisions for any $X \in \mathcal{X}$. (Right) The optimal decision rule $\delta^*(\cdot)$ satisfies $g\{X, \delta^*(X)\} = \inf_\delta g(X,\delta)$ for all $X \in \mathcal{X}$, and therefore minimizes $\int_{\mathcal{X}}g\{X,\delta(X)\}\intd\mu(X)$
under any strictly positive measure $\mu(\cdot)$ on $\mathcal{X}$. 
 \label{fig:Brown_Purves}}
\end{figure}

For illustration, consider the objective function 
$g(X,\delta) \equiv  \log\{(3 - X)^2 + 100(\delta - X^2)^2  + 1\}$; see Figure~\ref{fig:Brown_Purves}, left panel. 
 In a non-amortized setting, the optimal decision $\delta^*$ is found by minimizing $g(X,\delta)$ for fixed $X$. This potentially burdensome optimization problem must be repeated for each $X$. On the other hand, in an amortized setting, one finds a decision rule with a closed-form expression that returns an approximately optimal decision for any $X\in\mathcal{X}$. For our choice of $g(X,\delta)$ we have an exact closed-form expression for the optimal decision rule:  $\delta^*(X) = X^2$; see Figure~\ref{fig:Brown_Purves}, centre panel. 
 Furthermore, since $g\{X, \delta(X)\}$ is minimized pointwise by $\delta^*(\cdot)$ for each $X \in \mathcal{X}$ (see Figure~\ref{fig:Brown_Purves}, right panel) and $g(\cdot,\cdot)$ is nonnegative, the optimal decision rule satisfies $\delta^*(\cdot) = \argmin_{\delta(\cdot)} \int_{\mathcal{X}}g\{X,\delta(X)\}\intd\mu(X)$ under any strictly positive measure $\mu(\cdot)$ on $\mathcal{X}$. This property, which we refer to as \emph{average optimality}, is used for finding approximations to optimal decision rules.

A ubiquitous choice for $g(\cdot,\cdot)$ is a Kullback--Leibler (KL) divergence  \citep{Kullback_1951} which involves a target distribution $p(\cdot)$ and an approximate distribution $q(\cdot)$. For example, in posterior inference (Section~\ref{sec:neural_posterior_inference}), $p(\cdot)$ is a posterior distribution, $q(\cdot)$ is an approximate posterior distribution, and the decision rule $\deltab^*(\cdot)$ returns parameters of $q(\cdot)$ from data $\Zvec$. In the context of likelihood-function  approximation (Section~\ref{sec:NLE}), $p(\cdot)$ is a likelihood function, $q(\cdot)$ is an approximate (or synthetic) likelihood function, and $\deltab^*(\cdot)$ returns parameters used to construct $q(\cdot)$ from model parameters $\thetab$.  In both contexts, amortization is achieved by constructing approximate closed-form representations of optimal decision rules; we focus on cases where neural networks are used for this approximation.

\begin{marginnote}
  \entry{Kullback--Leibler (KL) divergence}{A dissimilarity measure between two distributions. For two densities $p(\cdot)$ and $q(\cdot)$, the forward KL divergence between $p(\cdot)$ and $q(\cdot)$ is $\KL{p(\Xvec)}{q(\Xvec)} \equiv \int p(\Xvec) \log \frac{p(\Xvec)}{q(\Xvec)} \intd \Xvec$. We denote the reverse KL divergence by $\KL{q(\Xvec)}{p(\Xvec)}$.}
    \end{marginnote}

\section{NEURAL POSTERIOR INFERENCE}\label{sec:neural_posterior_inference}

 We now review two prominent classes of amortized inferential approaches: neural Bayes estimation (Section~\ref{sec:neural_point_estimation}) and methods for neural posterior inference (Section~\ref{sec:KLmethods}) that approximate Bayesian posterior distributions via the minimization of an expected KL divergence. 

 \subsection{Neural Bayes Estimators}\label{sec:neural_point_estimation}

 Here we consider the decision-theoretic framework of Section~\ref{sec:Brown} where $\Xvec$ is data $\Zvec \in \mathcal{Z}$, the decision $\deltab$ is an estimate $\thetab \in \Theta$, and the decision rule $\deltab(\cdot)$ is an estimator $\thetabhat: \mathcal{Z} \rightarrow \Theta$. Consider a loss function $L: \Theta\times\Theta\rightarrow \mathbb{R}^{\ge0} $ and let $g\{\Zvec, \thetabhat(\Zvec)\} =\int_\Theta L\{\thetab,\widehat{\thetab}(\Zvec)\}p(\thetab \mid \Zvec)\intd \thetab$, the posterior expected loss. Then, by average optimality,
\begin{align}
\thetabhat^*(\cdot) &=  \argmin_{\thetabhat(\cdot)} \mathbb{E}_{\Zvec}\left[\int_\Theta L\{\thetab,\widehat{\thetab}(\Zvec)\}p(\thetab \mid \Zvec)\intd\thetab\right].\label{eq:forwardKL}
\end{align}
An application of Fubini's theorem to Equation~\ref{eq:forwardKL} \citep[][Theorem 2.3.2]{Robert_Choice} and Bayes' rule reveals that $\thetabhat^*(\cdot) =  \argmin_{\thetabhat(\cdot)} r_\textrm{B}\{\thetabhat(\cdot)\}$ where 
\begin{equation}\label{eq:BayesRisk}
  r_\textrm{B}\{\widehat{\thetab}(\cdot)\} \equiv\int_\Theta\int_\mathcal{Z} L\{\thetab,\widehat{\thetab}(\Zvec)\}p(\Zvec\mid \thetab)p(\thetab)\intd\Zvec\intd\thetab,
\end{equation}
is the Bayes risk. A minimizer of $r_\textrm{B}\{\widehat{\boldsymbol{\theta}}(\cdot)\}$ is said to be a Bayes estimator. Equation~\ref{eq:BayesRisk} can also be derived by considering the approximate posterior distribution $q\{\thetab; \thetabhat(\Zvec)\} \propto \exp[-L\{\thetab; \thetabhat(\Zvec)\}]$ and setting  $g\{\Zvec, \thetabhat(\Zvec)\} = \textrm{KL}[p(\thetab \mid \Zvec) ~\|~ q\{\thetab; \thetabhat(\Zvec)\}]$,  the forward KL divergence between ${p(\,\cdot \mid \Zvec)}$ and $q\{\,\cdot\,; \thetabhat(\Zvec)\}$. Bayes estimators are thus a special case of the forward KL divergence minimization methods reviewed in Section~\ref{sec:forwardKL}. 

\begin{figure}[t!]
  \begin{center}

  \begin{tikzpicture}[>=latex]

    \node[circle, draw, minimum size=1cm, inner sep=0pt] (Z) at (0,0) {$\Zvec$};
    \node[rectangle, draw, minimum size=1cm, inner sep=0pt, fill=black, text=white] (NN) at (3,0) {\textbf{NN}};
    \node[circle, draw, minimum size=1cm, inner sep=0pt] (thetabhat) at (6,0) {$\thetabhat$};

    \draw[->] (Z) to (NN);
    \draw[->] (NN) to (thetabhat);
    
    \draw[decorate,decoration={brace,amplitude=10pt,mirror}] 
([shift={(-0.2,-0.3)}]Z.south west) -- ([shift={(0.2,-0.3)}]Z.south east) 
node[midway,below=10pt] {Data};

    \draw[decorate,decoration={brace,amplitude=10pt,mirror}] 
([shift={(-0.2,-0.3)}]NN.south west) -- ([shift={(0.2,-0.3)}]NN.south east) 
node[midway,below=10pt,align=center] {Neural Bayes \\ Estimator};

\draw[decorate,decoration={brace,amplitude=10pt,mirror}] 
([shift={(-0.2,-0.3)}]thetabhat.south west) -- ([shift={(0.2,-0.3)}]thetabhat.south east) 
node[midway,below=10pt] {Point Estimate};

      \end{tikzpicture}
    \end{center}
\caption{Graphical representation of the neural Bayes estimator that takes data as input and that outputs a parameter point estimate.\label{fig:NBE}}
  \end{figure}
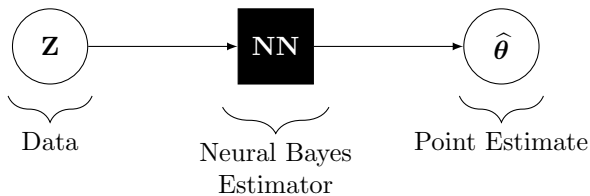

Assume now that  $\widehat{\thetab}_{\gammab}(\cdot)$ is a neural network flexible enough to approximate the true Bayes estimator arbitrarily well, where $\gammab$ are the network's parameters. Its architecture is largely determined by the structure of the data: For example, if the data are images, then $\thetabhat_{\gammab}(\cdot)$ is typically a convolutional neural network (CNN), which is designed for gridded data.  A simulation-based approach to train the network proceeds as follows: In Equation~\ref{eq:BayesRisk}, replace the estimator with the neural network; approximate the Bayes risk with a Monte Carlo approximation using $N$ simulations of parameters and corresponding datasets from an underlying model, $\{\{\thetab^{(i)}, \Zvec^{(i)}\}:i = 1,\dots,N\}$; 
and then solve the empirical risk minimization problem \begin{marginnote}
  \entry{Loss function}{A nonnegative function $L(\thetab, \thetabhat)$  which, in point estimation, quantifies the loss incurred when using $\thetabhat$ as an estimate of $\thetab$.}
  \entry{Neural-network architecture}{The specific design of a neural network, including the operations executed in each layer, how the layers are connected, and the number of layers.}
  \entry{Convolutional neural network (CNN)}{A neural network architecture involving multiple convolutional operations using kernels that are estimated during training.}
\end{marginnote}
\begin{equation}\label{eq:approxrisk}
 \gammab^* = \argmin_\gammab \sum_{i=1}^N L\{\thetab^{(i)},
 \thetabhat_{\gammab}(\Zvec^{(i)})\}.
\end{equation} 
Minimization is typically done using stochastic gradient descent in conjunction with automatic differentiation, implemented using machine-learning software libraries (see Section~\ref{sec:software}). We call the trained network a neural Bayes estimator. See \citet[][Section 2]{Sainsbury_2024} for more discussion and for an example comparing a neural Bayes estimator to the analytic Bayes estimator for a simple model. A graphical representation of the neural Bayes estimator is given in Figure~\ref{fig:NBE}.

Neural networks are extensively used for parameter point estimation. \citet{Gerber_2021}, \citet{Lenzi_2023}, and \citet{Sainsbury_2024} use CNNs to estimate covariance function parameters in spatial Gaussian process models and models of spatial extremes that are considered computationally difficult to fit. \citet{Liu_2020} also estimate the parameters of a Gaussian process, but they adopt a transformer network to cater for realizations of varying dimension and highly parameterized covariance functions. 
 \citet{Zammit_2020} use a CNN with three-dimensional kernels to estimate the dynamical parameters of an advection-diffusion equation, while \citet{Rudi_2022} apply a CNN with one-dimensional kernels for parameter estimation with ordinary differential equations. In all cases, a few minutes to a few hours are needed to train the neural networks, but once trained they provide estimates at a small fraction of the time required by classical estimators that require optimization or MCMC for any $\Zvec \in \mathcal{Z}$ (a speedup of at least a hundredfold is typical). A side benefit 
of quick parameter estimation is the quantification of uncertainty via parametric or non-parametric bootstrap techniques that require minimal extra computation.

Bayes estimators are functionals of the posterior distribution: Under the squared-error loss $L_{\textrm{se}}\{\thetab, \widehat\thetab(\Zvec)\} \equiv \|\widehat\thetab(\Zvec) - \thetab\|^2$, the Bayes estimator is the posterior mean $\E(\thetab \mid \Zvec)$; under the loss ${L_{\textrm{var}}\{\theta_j, \thetahat_j(\Zvec)\} \equiv [\{\theta_j - \E(\theta_j \mid\Zvec)\}^2 - \thetahat_j(\Zvec)]^2}$ for some parameter $\theta_j, j = 1,\dots,d$, the Bayes estimator is the posterior variance $\mathbb{V}(\theta_j\mid \Zvec)$ \citep{Fan_1998}; and under the quantile loss function
$ L_\rho\{\theta_j, \thetahat_j(\Zvec)\} \equiv  \{\thetahat_j(\Zvec) - \theta_j\}[\mathbbm{1}\{\thetahat_j(\Zvec) > \theta_j\} - \rho]$
the Bayes estimator is the posterior $\rho$-quantile. Posterior quantiles can be used for fast quantification of posterior uncertainty of $\thetab$ via the construction of credible intervals. For example, \citet{Sainsbury_2023} obtain posterior medians and 95\% credible intervals of Gaussian-process-covariance-function parameters in over 2,000 spatial regions from a million observations of sea-surface temperature in just three minutes on a single GPU. 
The required time is much less than that required by classical techniques such as non-amortized MCMC or variational inference.

\begin{marginnote}
  \entry{Amortization gap}{Error introduced in amortized inference because of incomplete training (e.g., not enough simulations), or because of neural network inflexibility, or both.}
  \end{marginnote}
  
  The outer expectation in Equation~\ref{eq:forwardKL} leads to the point estimator $\thetabhat^*(\cdot)$ being optimal, in a Bayes sense, for any $\Zvec$ (see Section~\ref{sec:Brown}). However, in practice, there will be some discrepancy between a trained neural network $\thetabhat_{\gammab^*}(\cdot)$ and the true Bayes estimator $\thetabhat^*(\cdot)$.  Any extra bias or variance introduced that makes the trained estimator sub-optimal with respect to the target optimal (in a KL sense) estimator is referred to as the amortization gap \citep{Cremer_2018}, and this gap is a consideration for this and all the remaining approaches discussed in this review.

\subsection{Approximate Bayesian inference via KL-divergence minimization}\label{sec:KLmethods}

Minimizing the KL divergence between $p(\thetab \mid \Zvec)$ and an approximate posterior distribution underpins many inference techniques. We focus on amortized versions of the two main techniques: forward KL divergence minimization, and reverse KL divergence minimization (i.e., variational Bayes). These approximate Bayesian methods are often used when asymptotically exact methods such as MCMC are computationally infeasible. The approximate distribution $q(\thetab; \kappab)$ has parameters $\kappab$ that need to be estimated. For example, when $q(\thetab\,;\kappab)$ is chosen to be Gaussian, the parameters $\kappab = (\mub', \vech(\Lmat)')'$ comprise a $d$-dimensional mean parameter $\mub$ and the $d(d+1)/2$ non-zero elements of the lower Cholesky factor $\Lmat$ of a covariance matrix, where the half-vectorization operator $\vech(\cdot)$ vectorizes the lower triangular part of its matrix argument. 

\begin{marginnote}
  \entry{Variational Bayes}{A type of approximate Bayesian inference where the approximate distribution is that which minimizes the reverse KL divergence between the true posterior distribution and its approximation.}
 \end{marginnote}
Forward and reverse KL minimization approaches both target the true posterior distribution, in the sense that in both cases the KL divergence is zero if and only if $q(\thetab)$ is identical to $p(\thetab \mid \Zvec)$ for $\thetab \in \Theta$. However, when $p(\thetab \mid\Zvec)$ is not in the class of approximating distributions, the optimal approximate distribution depends on the direction of the KL divergence. As shown by \citet[][Chapter~21]{Murphy_2012}, minimizing the reverse KL divergence leads to approximate distributions that are under-dispersed and that tend to concentrate mass on a single mode of the target distribution, whereas minimizing the forward KL divergence leads to ones that are over-dispersed and that cover all modes of the target distribution. 
Although both approaches are ubiquitous, in simulation-based settings 
forward KL approaches offer some advantages over their reverse KL counterparts; namely, they are natively likelihood-free and they have a more straightforward implementation.

\subsubsection{Forward KL-divergence minimization}\label{sec:forwardKL}

We first consider the non-amortized context, where we find the optimal approximate-distribution parameters $\kappab$ by forward KL divergence minimization:
\begin{align}
  \kappab^* &= \argmin_\kappab \KL{p(\thetab \mid \Zvec)}{q(\thetab ;\kappab)}\label{eq:fKL_optim}.
\end{align}
The optimization problem in Equation~\ref{eq:fKL_optim} is typically computationally expensive to solve even for a single $\Zvec$: Solving it for many different datasets is often computationally prohibitive. The learning problem can be amortized as described in Section~\ref{sec:Brown}, by treating the decision $\deltab$ as the parameter vector $\kappab$ and the decision rule $\deltab(\cdot)$ as the function $\kappab(\cdot)$ that maps data to distribution parameters. We then replace $q(\thetab; \kappab)$ with $q\{\thetab\,; \kappab(\Zvec)\}$ for $\thetab \in \Theta, \Zvec \in \mathcal{Z}$, set $g\{\Zvec,\kappab(\Zvec)\}$ to be the forward KL divergence between $p(\cdot)$ and $q(\cdot)$, and use average optimality to assert that
\begin{equation}
\label{eq:ForKLopt}
\kappab^*(\cdot) = \argmin_{\kappab(\cdot)} \mathbb{E}_{\Zvec}(\KLb{p(\thetab \mid \Zvec)}{q\{\thetab; \kappab(\Zvec)\}}).
\end{equation}

\begin{marginnote}
\entry{Inference network}{A neural network whose output can be used to construct an approximate posterior distribution from data $\Zvec$ or summaries thereof.}
\end{marginnote}

\begin{figure}[t!]

  \begin{center}

  \begin{tikzpicture}[>=latex]

    \node[circle, draw, minimum size=1cm, inner sep=0pt] (Z) at (0,0) {$\Zvec$};
    \node[rectangle, draw, minimum size=1cm, inner sep=0pt, fill=black, text=white] (NN) at (3,0) {\textbf{NN}};
    \node[circle, draw, minimum size=1cm, inner sep=0pt] (kappab) at (6,0) 
    {$\kappab$};

    \draw[->] (Z) to (NN);
    \draw[->] (NN) to (kappab);
    \draw[dashed,->] (kappab);

    \draw[decorate,decoration={brace,amplitude=10pt,mirror}] 
([shift={(-0.2,-0.3)}]Z.south west) -- ([shift={(0.2,-0.3)}]Z.south east) 
node[midway,below=10pt] {Data};

    \draw[decorate,decoration={brace,amplitude=10pt,mirror}] 
([shift={(-0.2,-0.3)}]NN.south west) -- ([shift={(0.2,-0.3)}]NN.south east) 
node[midway,below=10pt,align=center] {Inference \\ Network};

    \draw[decorate,decoration={brace,amplitude=10pt,mirror}] 
([shift={(-0.2,-0.3)}]kappab.south west) -- ([shift={(0.2,-0.3)}]kappab.south east) 
node[midway,below=10pt, align=center] {Approx.~Dist. \\ Parameters};

      \end{tikzpicture}
    \end{center}
    \caption{Graphical representation of an inference network that takes data as input and that outputs parameters of an approximate posterior distribution.\label{fig:InfNetwork}}
  \end{figure}
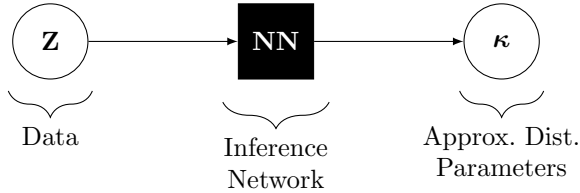

In neural amortized inference, one uses a neural network $\kappab_\gammab(\cdot)$ for $\kappab(\cdot)$, where $\gammab$ are the neural network parameters. The function $\kappab_\gammab(\cdot)$ is often referred to as an inference network, and  $\gammab$ is found by minimizing a Monte Carlo approximation to the expected KL divergence in Equation~\ref{eq:ForKLopt}. Specifically, the optimization problem becomes 
 \begin{align}\label{eq:MPopt}
      \gammab^* &= \argmin_{\gammab} -\sum_{i=1}^N \log q\{\thetab^{(i)}; \kappab_\gammab(\Zvec^{(i)})\},
    \end{align}
    where $\thetab^{(i)}\sim p(\thetab)$ and $\Zvec^{(i)} \sim p(\Zvec \mid \thetab^{(i)})$. This approach to neural posterior inference has the graphical representation shown in Figure~\ref{fig:InfNetwork}.

\citet{Chan_2018} let $q\{\cdot\,; \kappab_{\gammab}(\Zvec)\}$ be Gaussian with mean and precision both functions of the data $\Zvec \in \mathcal{Z}$, while \citet{Maceda_2024} consider a trans-Gaussian approximate posterior distribution.  \citet{Papamakarios_2016} go further and let $q\{\cdot\,;\kappab_{\gammab}(\Zvec)\}$ be a Gaussian mixture. Further flexibility can be achieved by modeling $q\{\cdot\,; \kappab_{\gammab}(\Zvec)\}$ using a normalizing flow. In such cases, the architecture used to construct $q\{\cdot\,; \kappab_\gammab(\Zvec)\}$ is often termed a conditional invertible neural network, since the invertible map constructed using the neural network has parameters determined by $\Zvec$; see, for example, \citet{Ardizzone_2018} 
    and \citet{Radev_2022}.\begin{marginnote}
\entry{Normalizing flow}{A sequence of invertible mappings used to establish a flexible family of distributions.}
\entry{Invertible neural network}{A neural network whose architecture constrains it to lie in the space of invertible mappings.}
\end{marginnote}From Equation~\ref{eq:MPopt} note that we must be able to evaluate $q\{\thetab; \kappab(\Zvec)\}$ for any $\thetab \in \Theta$ and $\Zvec \in \mathcal{Z}$ when training the neural network. This is possible using normalizing flows, which are invertible by construction. 
    However, invertibility can be a restrictive requirement; \citet{Pacchiardi_2022_GANs_scoring_rules} present an approach to circumvent this restriction by instead using a proper scoring rule \citep{Gneiting_2007_scoring_rules} in the objective function (Equation~\ref{eq:MPopt}).

\subsubsection{Reverse KL-divergence minimization}\label{sec:NVI}

Several methods for amortized inference minimize the expected reverse KL divergence between the true posterior distribution and the approximate posterior distribution, 
$$
\kappab^*(\cdot) = \argmin_{\kappab(\cdot)} \mathbb{E}_{\Zvec}(\KLb{q\{\thetab; \kappab(\Zvec)\}}{p(\thetab \mid \Zvec)}).
$$
In this case, the approximate posterior distribution is referred to as a variational posterior distribution. As in Section~\ref{sec:forwardKL}, the parameters appearing in the variational posterior distribution are functions of the data. The decision rule $\kappab^*(\cdot)$ minimizes a KL divergence between the true posterior distribution $p(\thetab \mid \Zvec)$ and the approximate posterior distribution for every $\Zvec \in \mathcal{Z}$, and hence yields optimal variational-posterior-distribution parameters for every $\Zvec \in \mathcal{Z}$. We now replace $\kappab(\cdot)$ with an inference (neural) network $\kappab_\gammab(\cdot)$ \citep{Mnih_2014}. The neural network parameters $\gammab$ can be optimized by minimizing a Monte Carlo approximation to the expected reverse KL divergence,

\begin{equation}\label{eq:VB_optim3}
\gammab^* \approx \argmin_{\gammab} \sum_{k = 1}^K\sum_{j = 1}^J  \left[\log q\{\thetab^{(j)}; \kappab_\gammab(\Zvec^{(k)})\} - \log\{p(\Zvec^{(k)} \mid \thetab^{(j)})p(\thetab^{(j)})\}\right], 
\end{equation}
where $\Zvec^{(k)} \sim p(\Zvec)$, and $\thetab^{(j)} \sim q\{\thetab; \kappab_\gammab(\Zvec^{(k)})\}$. The objective function in Equation~\ref{eq:VB_optim3} involves samples from both the model and the variational distribution. Since the latter quantity is itself a function of $\gammab$, one generally invokes the so-called reparameterization trick on $q\{\thetab; \kappab_\gammab(\Zvec^{(k)})\}$ to ensure that gradients with respect to $\gammab$ can be evaluated \citep{Kingma_2013}, which are required for training the neural network.
\begin{marginnote}
    \entry{Reparameterization trick}{Implicitly defining a parameter-dependent distribution via a generative model through which gradients can be propagated. For example, re-expressing $X \sim \mathrm{Gau}(\mu,\sigma^2)$ as ${X = \mu + \sigma\epsilon},$ ${\epsilon \sim \mathrm{Gau}(0, 1)}$.}
\end{marginnote}

The concept of an inference network that returns parameters of a variational posterior distribution dates back to \citet{Dayan_1995}, where $\kappab_{\gammab}(\cdot)$ was called a recognition model. It has since been used in various ways, for example for inference on hyperparameters in Gaussian process models \citep{Rehn_2022} or on latent variables within deep Gaussian process models \citep{Dai_2015}. \citet{Rezende_2015} and \citet{Kingma_2016} add flexibility to the variational posterior distribution by constructing $q\{\,\cdot\,; \kappab_\gammab(\Zvec)\}, \Zvec \in \mathcal{Z},$ using a normalizing flow, so that the inference network $\kappab_\gammab(\cdot)$ returns both the parameters of the flow's starting distribution and those governing the flow. 
Variational auto-encoders \citep{Kingma_2013} comprise an inference network in the encoding stage that maps the data directly to parameters of the approximate posterior distribution. More in-depth treatments of amortized variational inference are provided by \citet{Zhang_2018}, \citet{Ganguly_2023}, and \citet{Margossian_2023}.

Despite their appeal, variational learning networks are limited by the fact that Equation~\ref{eq:VB_optim3} involves the likelihood function $p(\Zvec \mid \thetab)$. 
 In some cases, this function is known and tractable; for example, Equation~\ref{eq:VB_optim3} can be used to develop an amortized variational inference framework for solving inverse problems \citep{Goh_2019, Svendsen_2023}, in which the conditional distribution of $\Zvec$ is assumed to be Gaussian with mean equal to the output of a forward physics model that takes $\thetab$ as input. \citet{Zhang_2023} also adopt amortized variational inference for estimating parameters in a hierarchical model for spatial extremes that has a tractable likelihood function. When the likelihood function is intractable, however, reverse KL-minimization techniques often use neural-network approaches for likelihood approximation; see Section~\ref{sec:NLE} for further details.

\section{NEURAL SUMMARY STATISTICS}\label{sec:summarystatistics}

Simulation-based inferential approaches often require pre-specified summary statistics, provided through expert judgement, indirect inference \citep[e.g.,][]{Drovandi_2011} or point summaries of the posterior distribution \citep[e.g.,][]{Fearnhead_2012}. \citet{Blum_2010}, \citet{Creel_2017}, \citet{Gerber_2021}, and \cite{Rai_2023}, for example, train neural Bayes estimators using hand-crafted summary statistics rather than raw data.  
In Section~\ref{sec:explicit_summ}, we review methods that employ neural networks to explicitly construct summary statistics for downstream inference; in Section~\ref{sec:implicit_summ}, we discuss methods that integrate summary-statistic construction implicitly in an amortized inference framework; and, in Section~\ref{sec:num_summ}, we discuss how the number of summary statistics might be chosen.
\begin{marginnote} 
  \entry{Indirect inference}{Fitting an auxiliary model, that is simpler than the target model, to data, and making inference on target parameters from the fitted auxiliary parameters.}
\end{marginnote}

 \subsection{Explicit neural summary statistics}\label{sec:explicit_summ}
 
 One of the most common neural summary statistics is the neural Bayes estimator. \citet{Jiang_2017}, for example, use a neural Bayes estimator under squared error loss to obtain the posterior mean quickly for use in ABC, while \citet{Dinev_2018} use a neural Bayes estimator in conjunction with the ratio-based likelihood-free inference method of \citet{Thomas_2022_ratio_estimation}; similar approaches are also proposed by \citet{Creel_2017, Akesson_2021} and \citet{Albert_2022}.
Beyond point estimators, neural networks can be trained to return other summaries that are highly informative of $\thetab$. A common approach is to target sufficiency of the summary statistics $\Smat(\Zvec)$ by maximizing the mutual information between $\thetab \in \Theta$ and $\Svec(\Zmat) \in \mathcal{S}$, where $\mathcal{S}$ is the sample space of $\Svec(\Zmat)$ (see \citeauthor{Barnes_2011}, \citeyear{Barnes_2011}, and the discussion by Barnes, Filippi and Stumpf in \citeauthor{Fearnhead_2012}, \citeyear{Fearnhead_2012}). The mutual information, $\textrm{MI}\{\thetab; \Svec(\Zmat)\}$, is defined as the KL divergence between the joint distribution $p\{\thetab,\Smat(\Zvec)\}$ and the product of marginals distributions $p(\thetab)p\{\Smat(\Zvec)\}$. 
Intuitively, $\thetab$ and $\Smat(\Zvec)$ are nearly independent when the mutual information is small, so the summary statistic $\Smat(\Zvec)$ is irrelevant for describing or predicting $\thetab$, and conversely when the mutual information is large. Assume now that a neural network is used to construct summary statistics from data. The so-called summary network $\Smat_{\taub}(\cdot)$ with network parameters $\taub$ is trained by solving
\begin{marginnote}
    \entry{Sufficient statistic}{A statistic $\Svec(\Zvec)$ is sufficient for $\thetab$ if and only if $p(\Zvec\mid\thetab)=h(\Zvec)f\{\Svec(\Zvec);\thetab\}$ for some nonnegative functions $h(\cdot),f(\cdot)$ (Fisher--Neyman Factorization Theorem).}
\end{marginnote}
\begin{equation}\label{eq:MI-KL-max} \taub^* = \argmax_{\taub} \textrm{MI}\{\thetab; \Svec_{\taub}(\Zmat)\} = \argmax_{\taub} \KLb{p\{\thetab, \Svec_{\taub}(\Zvec)\}}{p(\thetab) p\{\Svec_{\taub}(\Zvec)\}}.\end{equation}
The objective in Equation~\ref{eq:MI-KL-max} is difficult to approximate using Monte Carlo techniques since it involves the distribution of the unknown summary statistics. To circumvent this, the mutual information neural estimator \citep[MINE;][]{Belghazi_2018} replaces the KL divergence with its Donsker--Varadhan (\citeyear{Donsker_1983}) representation. Writing $(\thetab',\Svec_{\taub}(\Zvec)')'\sim p\{\thetab, \Svec_{\taub}(\Zvec)\}$ and $(\tilde{\thetab}',{\Svec}_{\taub}(\tilde{\Zvec})')'\sim p(\thetab)p\{\Svec_{\taub}(\Zvec)\}$, we have
\begin{equation*}
\textrm{MI}\{\thetab; \Svec(\Zmat)\} = \sup_{T: \Theta\times \mathcal{S} \rightarrow \mathbb{R}}\E_{(\thetab', \Svec_{\taub}(\Zvec)')'}[T\{\thetab, \Svec_{\taub}(\Zvec)\}] - \log\E_{(\tilde{\thetab}', {\Svec}_{\taub}(\tilde{\Zvec})')'}[\exp T\{\tilde{\thetab}, {\Svec}_{\taub}(\tilde{\Zvec})\}].
\end{equation*}
The appeal of this representation is that it does not require the distribution of the summary statistics when approximated using Monte Carlo methods. In MINE, the function $T(\cdot,\cdot)$ is modeled using a neural network $T_{\zetab}(\cdot, \cdot)$ that is trained in tandem with $\Svec_{\taub}(\cdot)$ via
\begin{equation*}
 (\taub^{*'}, \zetab^{*'})' = \argmax_{(\taub', \zetab')'} \frac{1}{N}\sum_{i=1}^N T_{\zetab}\{\thetab^{(i)}, \Svec_{\taub}(\Zvec^{(i)})\} - \log\left(\frac{1}{N}\sum_{i=1}^N \exp[T_{\zetab}\{\thetab^{(i)}, \Svec_{\taub}(\Zvec^{\{\pi(i)\}})\}]\right),
\end{equation*}
where $\thetab^{(i)} \sim p(\thetab)$, $\Zvec^{(i)} \sim p(\Zvec \mid \thetab^{(i)})$ and $\pi(\cdot)$ is a random permutation function used to ensure that  $({\thetab^{(i)'}},{\Svec}_{\taub}(\Zvec^{\{\pi(i)\}})')'\sim p(\thetab)p\{\Svec_{\taub}(\Zvec)\}$.

\citet{Hjelm_2018} found that optimization routines were more stable when the Donsker--Varadhan objective function is replaced with one based on the Jensen--Shannon divergence:
$$
\textrm{MI}\{\thetab; \Svec(\Zmat)\} \approx \sup_{T: \Theta\times \mathcal{S} \rightarrow \mathbb{R}}\E_{(\thetab', \Svec_{\taub}(\Zvec)')'}(-\textrm{sp}[-T\{\thetab, \Svec_{\taub}(\Zvec)\}]) - \E_{(\tilde{\thetab}',{\Svec}_{\taub}(\tilde{\Zvec})')'}(\textrm{sp}[T\{\tilde{\thetab}, {\Svec}_{\taub}(\tilde{\Zvec})\}]),
$$
where $\textrm{sp}(z) \equiv \log\{1+\exp(z)\}$ is the softplus function. The Jensen--Shannon divergence MI estimator was employed to extract summary statistics for use with both classical simulation-based inference methods and neural based inference methods by \citet{Chen_2021}. 
\citet{Chen_2023} propose to further tame the optimization problem by using a slice technique that effectively breaks down the high-dimensional information objective into many smaller, lower-dimensional ones.

\begin{marginnote}
    \entry{Softplus function}{The function $\textrm{sp}(z) \equiv \log\{1+\exp(z)\}$ commonly used as a smooth approximation to the rectified linear unit, $\textrm{ReLU}(z)=\max(0,z)$. Asymptotically, $\textrm{sp}(z)\approx\exp(z)$ as $z\to-\infty$ and $\textrm{sp}(z)\approx z$ as $z\to\infty$.}
\end{marginnote}

Other neural-network-based approaches to find summary statistics include those of \citet{Charnock_2018} and \citet{deCastro_2019}, who target statistics that maximize the determinant of the Fisher information matrix. \citet{Pacchiardi_2022} take yet another approach and fit a general exponential family model in canonical form to simulated data. Their model comprises a summary statistic network $\Smat_{\taub}(\cdot)$ and a network for the canonical parameters that together form a so-called neural likelihood function. The networks are fitted by minimizing the Fisher divergence between the true and the neural likelihood functions. Although the neural likelihood function is inferred up to a normalizing constant, and can be used with certain MCMC algorithms like the exchange algorithm, the main appeal of the method is that the neural summary statistics extracted are the sufficient statistics of the best (in a Fisher divergence sense) exponential family approximation to the true likelihood function.\begin{marginnote}
  \entry{Fisher divergence}{For two densities $p(\cdot)$ and $q(\cdot)$, the Fisher divergence between $p(\cdot)$ and $q(\cdot)$ is $D_F\{p(\Xvec)\| q(\Xvec)\} \equiv \frac{1}{2}\int p(\Xvec) \|\nabla\log p(\Xvec) - \nabla\log q(\Xvec) \|^2 \intd \Xvec$.}
\end{marginnote}    

\subsection{Implicit neural summary statistics}\label{sec:implicit_summ}

The ability of neural networks to extract relevant summary statistics from  data is analogous to their ability to learn features that are useful for predicting an outcome, for example in image recognition; this property is often referred to as representation learning. Feature extraction happens in the early layers of neural networks; similarly one can design networks for amortized inference so that the first few layers extract informative summary statistics.

\begin{marginnote}
    \entry{Representation learning}{The learning of features or summaries from data that are useful for downstream inference tasks.}
\end{marginnote}

\begin{figure}[t!]

  \begin{center}

  \begin{tikzpicture}[>=latex]

    \node[circle, draw, minimum size=1cm, inner sep=0pt] (Z) at (0,0) {$\Zvec$};
    \node[rectangle, draw, minimum size=1cm, inner sep=0pt, fill=black, text=white] (NN1) at (2.8,0) {\textbf{NN}};
    \node[circle, draw, minimum size=1cm, inner sep=0pt] (S) at (5.6,0) {$\Svec$};

    \node[rectangle, draw, minimum size=1cm, inner sep=0pt, fill=black, text=white] (NN2) at (8.4,0) {\textbf{NN}};
    \node[circle, draw, minimum size=1cm, inner sep=0pt] (kappab) at (11.2,0) 
    {$\kappab$};

    \draw[->] (Z) to (NN1);
    \draw[->] (NN1) to (S);
    \draw[->] (S) to (NN2);
    \draw[->] (NN2) to (kappab);
    \draw[dashed,->] (kappab);

    \draw[decorate,decoration={brace,amplitude=10pt,mirror}] 
([shift={(-0.2,-0.3)}]Z.south west) -- ([shift={(0.2,-0.3)}]Z.south east) 
node[midway,below=10pt] {Data};

    \draw[decorate,decoration={brace,amplitude=10pt,mirror}] 
([shift={(-0.2,-0.3)}]NN1.south west) -- ([shift={(0.2,-0.3)}]NN1.south east) 
node[midway,below=10pt,align=center] {Summary \\ Network};

    \draw[decorate,decoration={brace,amplitude=10pt,mirror}] 
([shift={(-0.2,-0.3)}]S.south west) -- ([shift={(0.2,-0.3)}]S.south east) 
node[midway,below=10pt,align=center] {Summary \\ Statistics};

    \draw[decorate,decoration={brace,amplitude=10pt,mirror}] 
([shift={(-0.2,-0.3)}]NN2.south west) -- ([shift={(0.2,-0.3)}]NN2.south east) 
node[midway,below=10pt,align=center] {Inference \\ Network};

    \draw[decorate,decoration={brace,amplitude=10pt,mirror}] 
([shift={(-0.2,-0.3)}]kappab.south west) -- ([shift={(0.2,-0.3)}]kappab.south east) 
node[midway,below=10pt, align = center] {Approx.~Dist. \\ Parameters};

      \end{tikzpicture}
    \end{center}
    \caption{Graphical representation depicting a summary network that takes data as input and that outputs summary statistics, and an inference network that takes the summary statistics as input and that outputs parameters of an approximate posterior distribution.\label{fig:SummInfNetwork}}
  \end{figure}
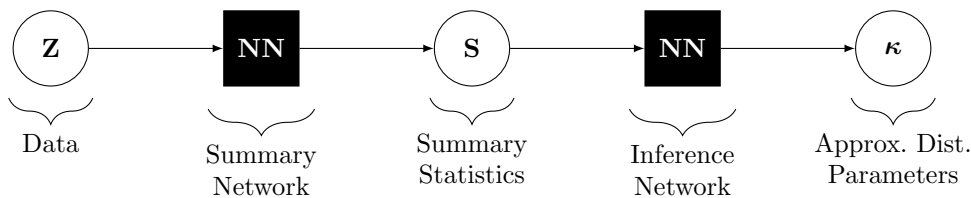

All methods discussed in Section~\ref{sec:neural_posterior_inference} can incorporate this design by splitting the underlying neural-network architecture into two parts: a summary network for extracting summary statistics from the data, and an inference network that uses these summary statistics as input to output parameter estimates or distributional parameters. The summary network can be trained in tandem with the now-simplified inference network in an end-to-end manner;  see, for example, \citet{Sainsbury_2024} for this setup in an estimation setting, \citet{Radev_2022} in a forward KL divergence setting, and \citet{Wiqvist_2021} in a reverse KL divergence setting.  The inference network combined with a summary network in an approximate Bayes setting has the graphical representation shown in Figure~\ref{fig:SummInfNetwork}. When a set of hand-crafted summary statistics is known to be informative of the parameter vector, one may use both these and implicitly learned (neural) summary statistics simultaneously \citep[see, e.g.,][Sec.~2.2.1]{Sainsbury_2024}.

\subsection{Number of summary statistics}\label{sec:num_summ}

The dimensionality of $\Svec(\cdot)$, that is, the number of summary statistics, is a design choice. \citet{Pacchiardi_2022} set the number equal to the number of model parameters while \citet{Chen_2021} make it twice as large. \citet{Chen_2023} propose letting $\Svec(\cdot)$ be reasonably high-dimensional and retaining summary statistics that substantially contribute to mutual information, noting that twice the dimensionality of $\thetab$ suffices in most cases. In the context of neural Bayes estimators, \citet{Gerber_2021} and \citet{Sainsbury_2024} set the number of (implicitly defined) summary statistics to 128, far exceeding the number of model parameters, but irrelevant statistics are downweighted during training.  In practice some experimentation might be needed to determine a suitable number of summary statistics, which should be sufficiently small for training purposes, but large enough for the amortization gap to be within a tolerable range for a given application.

\section{NEURAL LIKELIHOOD AND LIKELIHOOD-TO-EVIDENCE RATIO}\label{sec:likelihood_and_ratio}

Section~\ref{sec:neural_posterior_inference} describes methods for amortized neural posterior inference, while Section~\ref{sec:summarystatistics} outlines the role of summary statistics and how they can be automatically constructed from data. In this section we discuss methods for amortized approximation of the likelihood function, $p(\Zvec \mid \thetab)$ (Section~\ref{sec:NLE}), and a closely related quantity that is proportional to the likelihood function, the likelihood-to-evidence ratio, $r(\thetab, \Zvec)= p(\Zvec \mid \thetab)/p(\Zvec)$ (Section~\ref{sec:NRE}). 

Amortized neural likelihood and likelihood-to-evidence-ratio approximators have several common motivations. First, the likelihood function is the cornerstone of frequentist inference, while likelihood ratios of the form $p(\Zvec\mid \thetab_0)/p(\Zvec\mid \thetab_1)=r(\thetab_0, \Zvec)/r(\thetab_1, \Zvec)$ are central to hypothesis testing, model comparison, and naturally appear in the transition probabilities of most standard MCMC algorithms used for Bayesian inference. Second, amortized likelihood and likelihood-to-evidence-ratio approximators enable the straightforward treatment of conditionally (on $\thetab$) independent and identically distributed (i.i.d.) replicates since then one has $p(\Zvec_1, \dots, \Zvec_m \mid \thetab) = \prod_{i=1}^m p(\Zvec_i \mid \thetab)$, and the multiple-replicate likelihood-to-evidence ratio is of the form 
$p(\Zvec_1, \dots, \Zvec_m \mid \thetab) / p(\Zvec_1, \dots, \Zvec_m)
\propto \prod_{i=1}^m r(\Zvec_i, \thetab).$ 
Hence, an amortized likelihood or likelihood-to-evidence-ratio approximator constructed with single-replicate datasets can be used  for frequentist or Bayesian inference with an arbitrary number of conditionally i.i.d.~replicates. Third, as these amortized approximators target prior-free quantities, they are ideal for Bayesian inference tasks that require multiple fits of the same model under different prior distributions. However, as we explain below, these methods still require the user to define a proposal distribution $\tilde{p}(\thetab)$ from which parameters will be sampled during the training stage, which determines the regions of the parameter space where the amortization gap will be lowest.

\subsection{Neural Likelihood}\label{sec:NLE}

We organize this section as follows: In Sections~\ref{sec:NVI-synth} and \ref{sec:amortized_likelihood} we discuss neural synthetic likelihood and neural full likelihood, respectively; in Section~\ref{sec:proposal} we provide some comments on the proposal distribution used for training the neural networks; and in Section~\ref{sec:JANA} we outline an approach that implements neural likelihood and amortized inference simultaneously.

\subsubsection{Neural synthetic likelihood}\label{sec:NVI-synth}

A popular method to approximate a likelihood function is the synthetic likelihood framework of \citet{Wood_2010}. In this framework, one replaces the likelihood function $p(\Zvec \mid \thetab)$ with a synthetic one of the form $q\{\Smat(\Zvec);\omegab(\thetab)\}$  based on summary statistics $\Smat(\Zvec)$, where $\omegab(\thetab)$ is a binding function linking the parameter vector $\thetab$ to the (approximate) distribution of $\Svec(\Zvec)$  \citep[e.g.,][]{Moores_2015}. Here $q\{\Smat(\Zvec);\omegab(\thetab)\}$ refers to a generic (approximate) distribution of $\Smat(\Zvec)$ evaluated at $\Smat(\Zvec)$ itself. The summary statistics $\Smat(\cdot)$ are often modeled as Gaussian with mean parameter $\mub(\thetab)$ and covariance matrix $\Sigmamat(\thetab)$. If the binding function $\omegab(\thetab)=\{\mub(\thetab),\Sigmamat(\thetab)\}$ is known, then synthetic likelihood is a form of amortized likelihood approximation, as it allows one to evaluate $p(\,\cdot \mid \thetab)$ quickly for any $\thetab$. When a neural network is used to model the binding function, we obtain a neural synthetic likelihood. In the Gaussian case, we let $q\{\Smat(\Zvec);\omegab_{\etab}(\thetab)\}$ be a Gaussian function with mean parameter $\muvec_{\etab}(\thetab)$ and covariance matrix $\Sigmamat_{\etab}(\thetab)$, where the neural binding function is $\omegab_{\etab}(\thetab)=\{\muvec_{\etab}(\thetab),\Sigmamat_{\etab}(\thetab)\}$, and $\etab$ are neural network parameters to be estimated. Often, the Gaussianity assumption for $\Svec(\Zvec)$ is deemed too restrictive, so \citet{Radev_2023} model $q\{\Smat(\Zvec);\omegab_{\etab}(\thetab)\}$ using normalizing flows. 

\begin{marginnote}
  \entry{Synthetic likelihood function}{The distribution of summary statistics (typically assumed to be Gaussian) with parameters that are a function of the underlying model parameters of interest. Often used as a replacement for an intractable likelihood function.}
  \entry{Binding function}{Here used to refer to a function mapping the parameters of interest to the parameters of the approximate (synthetic) likelihood function.}
\end{marginnote}

\begin{figure}[t!]

  \begin{center}

  \begin{tikzpicture}[>=latex]

    \node[circle, draw, minimum size=1cm, inner sep=0pt] (thetab) at (0,0) {$\thetab$};
    \node[rectangle, draw, minimum size=1cm, inner sep=0pt, fill=black, text=white] (NN) at (3,0) {\textbf{NN}};
    \node[circle, draw, minimum size=1cm, inner sep=0pt] (kappab) at (6,0) 
    {$\omegab$};

    \draw[->] (thetab) to (NN);
    \draw[->] (NN) to (kappab);
    \draw[dashed,->] (thetab);

    \draw[decorate,decoration={brace,amplitude=10pt,mirror}] 
([shift={(-0.2,-0.3)}]Z.south west) -- ([shift={(0.2,-0.3)}]Z.south east) 
node[midway,below=10pt] {Parameters};

    \draw[decorate,decoration={brace,amplitude=10pt,mirror}] 
([shift={(-0.2,-0.3)}]NN.south west) -- ([shift={(0.2,-0.3)}]NN.south east) 
node[midway,below=10pt,align=center] {Neural \\ Binding Function};

    \draw[decorate,decoration={brace,amplitude=10pt,mirror}] 
([shift={(-0.2,-0.3)}]kappab.south west) -- ([shift={(0.2,-0.3)}]kappab.south east) 
node[midway,below=10pt, align=center] {Binding Function \\ Parameters};

      \end{tikzpicture}
    \end{center}
  \caption{Graphical representation of a neural binding function, a neural network that takes parameters as input and that outputs binding function parameters used to construct a synthetic likelihood function.\label{fig:SynthLikNetwork}}
  \end{figure}
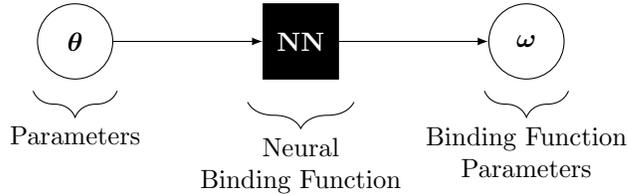

The binding function is a decision rule ($\deltab(\cdot)$ in Section~\ref{sec:Brown}) that maps the model parameters $\thetab$ to the parameters of the synthetic likelihood function. Choosing $g(\cdot,\cdot)$ to be the forward KL divergence between the true likelihood function and the synthetic one, we then apply average optimality to obtain a training objective for the neural binding function,
  \begin{equation}\label{eq:synthetic}
   \etab^* = \argmin_{\etab} \mathbb{E}_{\thetab}(\KLb{p(\Zvec \mid \thetab)}{q\{\Smat(\Zvec);\omegab_{\etab}(\thetab)\}}).
  \end{equation}
Equation \ref{eq:synthetic} leads us to consider the empirical risk minimization problem
  \begin{equation}  \label{eq:synthetic2}
  \etab^* = \argmin_{\etab} -\sum_{i=1}^N\log q\{\Smat(\Zvec^{(i)});\omegab_{\etab}(\thetab^{(i)})\},
  \end{equation}
where $\thetab^{(i)} \sim \tilde{p}(\thetab), i = 1,\dots,N,$ are drawn from some proposal distribution (further discussed below), and $\Zvec^{(i)} \sim p(\Zvec \mid \thetab^{(i)})$. The neural binding function has the  graphical representation shown in Figure~\ref{fig:SynthLikNetwork}.

  Once training is complete, the neural synthetic likelihood function can replace the true likelihood function in other inferential techniques, for example in amortized variational inference (Equation~\ref{eq:VB_optim3}), as suggested by \citet{Wiqvist_2021}. The resulting framework amortizes the conventional variational-inference-with-synthetic-likelihood approach of \citet[e.g.,][]{Ong_2018}; we demonstrate its use in Section~\ref{sec:software}.

\subsubsection{Neural full likelihood}
\label{sec:amortized_likelihood}

While Equation~\ref{eq:synthetic} leads to an amortized likelihood approximation framework, replacing the full likelihood with a synthetic counterpart may be undesirable (e.g., when one wishes to emulate the data $\Zvec$ for any given $\thetab$). One can instead target the full likelihood function, which is obtained by setting $\Smat(\cdot)$ to be the identity function. 
Equations~\ref{eq:synthetic}~and~\ref{eq:synthetic2} then become
\begin{align}
\etab^* &= \argmin_{\etab} \mathbb{E}_{\thetab}(\KLb{p(\Zvec \mid \thetab)}{q\{\Zvec;\omegab_{\etab}(\thetab)\}}) = \argmin_{\etab}\mathbb{E}_{(\thetab',\Zvec')'}[-\log q\{\Zvec;\omegab_{\etab}(\thetab)\}],\label{eq:like1}
\end{align}
and
\begin{equation}
\label{eq:like2}
\etab^{*} = \argmin_{\etab}-\sum_{i = 1}^N \log q\{\Zvec^{(i)};\omegab_{\etab}(\thetab^{(i)})\},
\end{equation}
respectively. Such a framework was considered by \citet{Lueckmann_2017} \citep[using Gaussian mixture density networks; see][Chapter 5]{Bishop_1995} and \citet{Papamakarios_2019} (using masked autoregressive flows).

\begin{marginnote}
\entry{Gaussian mixture density network}{A Gaussian mixture conditional probability distribution, where the means, variances, and mixing coefficients are the outputs of neural networks that take the conditioning variable as input. }
\end{marginnote}

\subsubsection{Choice of proposal distribution}\label{sec:proposal} 
In theory we can use any proposal distribution $\tilde{p}(\thetab)$ that is supported over the parameter space to obtain samples for Equation~\ref{eq:synthetic2} or \ref{eq:like2}. However, in practice, the chosen proposal distribution is an important consideration because $\tilde{p}(\thetab)$ determines the region of the parameter space where the parameters $\{\thetab^{(i)}\}$ will be most densely sampled, and thus, where the likelihood approximation will be most accurate. 
 Therefore, the proposal distribution should be sufficiently vague to cover the plausible values of $\thetab$, even though, if used in a Bayesian context, it need not be the prior distribution $p(\thetab)$. 
\citet{Papamakarios_2019} develop a sequential training scheme aimed at improving simulation efficiency (and thus likelihood approximation accuracy) for specific data $\Zvec$.

\subsubsection{Joint amortization}\label{sec:JANA} \citet{Radev_2023} propose to simultaneously approximate both the posterior distribution and the likelihood function within a single training regime, using two jointly-trained normalizing flows. The method, dubbed jointly amortized neural approximation (JANA), involves extending Equations~\ref{eq:ForKLopt}~and~\ref{eq:like1} to the joint optimization problem, 
\begin{align}
(\gammab^{*'},\taub^{*'},\etab^{*'})'=\argmax_{(\gammab^{'},\taub^{'},\etab^{'})'}&\mathbb{E}_{(\thetab',\Zvec')'}(\log q[\thetab;\kappab_{\gammab}\{\Svec_{\taub}(\Zvec)\}]+\log q\{\Zvec;\omegab_{\etab}(\thetab)\})\nonumber\\
&-\lambda\cdot\left(\textrm{MMD}\left[p\{\Svec_{\taub}(\Zvec)\} ~\|~ \textrm{Gau}(\zerob, \Imat)\right]\right)^2,\qquad \lambda>0, \label{eq:objJANA}
\end{align} \begin{marginnote}
  \entry{Maximum mean discrepancy}{A dissimilarity measure between two distributions. For two densities $p(\cdot)$ and $q(\cdot)$ we denote the discrepancy between $p(\cdot)$ and $q(\cdot)$ as $\textrm{MMD}\{p(\Xvec) ~\|~ q(\Yvec)\}=\sup_{g(\cdot) \in \mathcal{G}}[\mathbb{E}_{\Xvec}\{g(\Xvec)\}-\mathbb{E}_{\Yvec}\{g(\Yvec)\}]$, for $\Xvec\sim p(\cdot)$, $\Yvec\sim q(\cdot)$ and a suitable class of real-valued functions $\mathcal{G}$ \citep{Gretton_2012}.}

\end{marginnote}
\hspace{-0.15in} where $q[\thetab;\kappab_{\gammab}\{\Svec_{\taub}(\Zvec)\}]$ approximates the posterior distribution ${p(\thetab\mid\Zvec)}$ through the summary network $\Svec_{\taub}(\Zvec)$, and where $q\{\Zvec;\omegab_{\etab}(\thetab)\}$ (or a synthetic likelihood version, based on $\Svec_{\taub}(\Zvec)$) approximates the likelihood function $p(\Zvec\mid\thetab)$; the rightmost term on the right-hand side of Equation~\ref{eq:objJANA} involves the maximum mean discrepancy ($\textrm{MMD}$) between the distribution of the summary statistics, $p\{\Svec_{\taub}(\Zvec)\}$, and the standard multivariate Gaussian distribution, $\textrm{Gau}(\zerob, \Imat)$ -- this serves as a penalty to impose some probabilistic structure on the summary space and allows for the detection of model misspecification \citep[for further details, see][]{Radev_2023}. 
Since JANA approximates both the (normalized) posterior distribution and the likelihood function, it also automatically yields an amortized approximation to the marginal likelihood $p(\Zvec)=p(\Zvec\mid\thetab)p(\thetab)/p(\thetab\mid\Zvec)$, which is key for Bayesian model comparison and selection. Moreover, it also allows one to measure out-of-sample posterior predictive performance via the expected log-predictive distribution. Another joint amortization method, coined the Simformer \citep{Glockler_2024}, uses a score-based diffusion model \citep{Song_2021} for the joint distribution of $\thetab$ and $\Zvec$, in such a way that samples from any of its conditionals can be easily generated.

\subsection{Neural Likelihood-to-Evidence Ratio}\label{sec:NRE}

In Section~\ref{sec:lik_ratio_general} we outline the general framework for approximating the likelihood-to-evidence ratio, while in Section~\ref{sec:lik-ratio-variants} we discuss variants that safeguard against overoptimistic inferences and that aim to improve computational efficiency.

\subsubsection{General framework}\label{sec:lik_ratio_general}

\begin{figure}[t!]
  \centering
  
  \includegraphics[width=\linewidth]{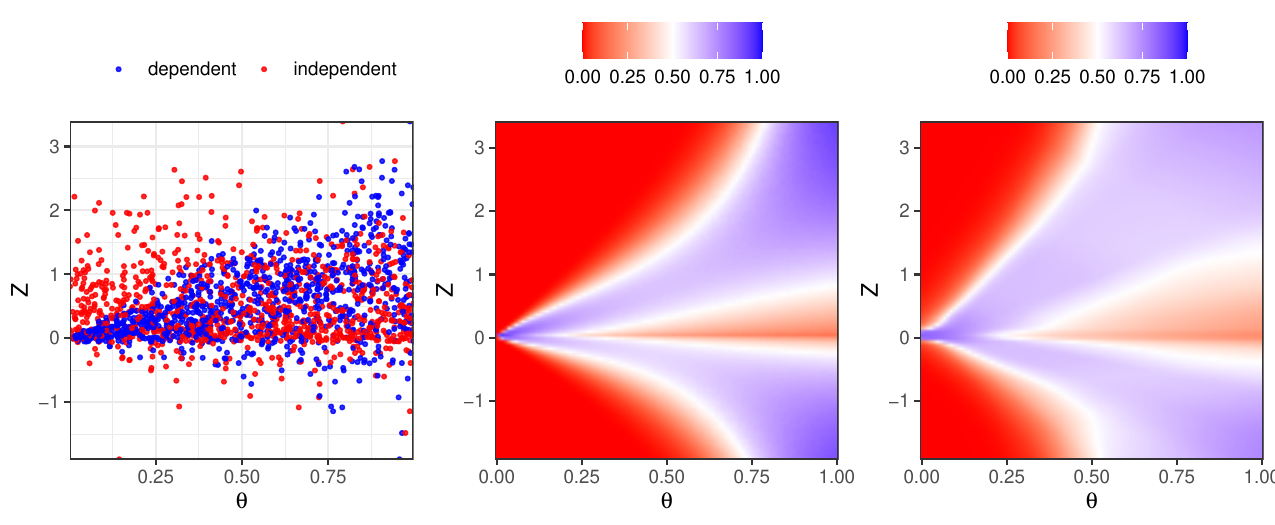}
  \caption{Illustration of amortized likelihood-to-evidence ratio approximation using the model $Z \mid \theta \sim \textrm{Gau}(\theta, \theta^2)$, $\theta \sim \textrm{Unif}(0, 1)$. (Left) Samples of dependent pairs $\{\theta, Z\} \sim p(\theta, Z)$ (blue) and independent pairs $\{\tilde{\theta}, \tilde{Z}\} \sim p(\theta)p(Z)$ (red). (Centre) Class probabilities from the exact Bayes classifier 
  $c^*(\theta, Z) = p(\theta, Z)\{p(\theta, Z) + p(\theta)p(Z)\}^{-1}$, linked to the likelihood-to-evidence ratio via the relation $r(\theta, Z) = c^*(\theta, Z)\{1 - c^*(\theta, Z)\}^{-1}$. 
  (Right) Class probabilities from an amortized neural approximation $c_{\gammab^*}(\cdot, \cdot)$ of the Bayes classifier.
  \label{fig:Bayes_classifier}}
  \end{figure}

Several approaches to likelihood-free inference are based on ratio approximation \citep{JMLR:gutmann12a}. Some approaches \citep[e.g.,][]{Cranmer_2015_ratio_estimators, Baldi_2016_NRE} target ratios of the form $p(\Zvec\mid\thetab)/p(\Zvec\mid\thetab_{\textrm{ref}})$ for some arbitrary but fixed reference parameter $\thetab_{\textrm{ref}}$. In this review we focus on a particular formulation that obviates the need for a reference parameter by instead targeting the likelihood-to-evidence ratio \citep{Hermans_2020, Thomas_2022_ratio_estimation},
\begin{equation}\label{eqn:likelihood-to-evidence_ratio}
    r(\thetab, \Zvec) = p(\Zvec \mid \thetab)/p(\Zvec). 
\end{equation}
This ratio can be approximated 
by solving a relatively straightforward binary classification problem, as we now show. Consider a binary classifier $c(\thetab, \Zvec)$ that distinguishes dependent parameter-data pairs ${(\thetab', \Zvec')' \sim p(\thetab, \Zvec \mid Y = 1) = p(\thetab, \Zvec)}$ from independent parameter-data pairs ${(\tilde{\thetab}', \tilde{\Zvec}')' \sim p(\thetab, \Zvec \mid Y = 0) = p(\thetab)p(\Zvec)}$, where $Y$ denotes the class label and where the classes are balanced. Then, we define the optimal classifier $c^*(\cdot,\cdot)$ as that which minimizes the Bayes risk under binary cross-entropy loss, $L_{\textrm{bce}}(\cdot,\cdot)$,
 \begin{align}
    c^*(\cdot, \cdot) 
    &\equiv \argmin_{c(\cdot, \cdot)} \sum_{y\in\{0, 1\}} \textrm{Pr}(Y = y) \int_\Theta\int_\mathcal{Z}p(\thetab, \Zvec \mid Y = y)L_{\textrm{bce}}\{y, c(\thetab, \Zvec)\}\intd \Zvec \intd \thetab \nonumber \\
    &= \argmax_{c(\cdot, \cdot)} \sum_{y\in\{0, 1\}} \int_\Theta\int_\mathcal{Z}p(\thetab, \Zvec \mid Y = y)[y\log\{c(\thetab, \Zvec)\} + (1 - y) \log\{1 - c(\thetab, \Zvec)\}]\intd \Zvec \intd \thetab \nonumber \\
    &= \argmax_{c(\cdot, \cdot)} \int_\Theta\int_\mathcal{Z}p(\thetab, \Zvec)\log \{c(\thetab, \Zvec)\} \intd \Zvec \intd \thetab + \int_\Theta\int_\mathcal{Z}p(\thetab)p(\Zvec)\log\{1 - c(\thetab, \Zvec)\}\intd \Zvec \intd \thetab \nonumber \\
    &= \argmax_{c(\cdot, \cdot)} \Big[\E_{(\thetab', \Zvec')'} \log \{c(\thetab, \Zvec)\}  +  \E_{(\tilde{\thetab}',\tilde{\Zvec}')'} \log \{1 - c(\tilde{\thetab}, \tilde{\Zvec})\}\Big], \label{eqn:optimal_classifier} 
\end{align}
where $\mathrm{Pr}(Y=y)$, $y\in\{0,1\}$, denotes the class probability. It can be shown that $c^*(\thetab, \Zvec) = p(\thetab, \Zvec)\{p(\thetab, \Zvec) + p(\thetab)p(\Zvec)\}^{-1}$ \citep[e.g.,][App.~B]{Hermans_2020} and, 
 hence, 
\begin{marginnote}
     \entry{Balanced classification task}{A classification task is said to be balanced if the prior class probabilities are equal.}
     \entry{Binary cross-entropy (log) loss}{The loss $L_{\textrm{bce}}(y, c) = -y\log(c) - (1 - y) \log(1 - c)$ (i.e., the negative Bernoulli log-likelihood), a measure of the difference between the true class $y \in \{0, 1\}$ and an estimated class probability $c \in (0, 1)$.}
\end{marginnote}
\begin{equation}\label{eqn:optimal_classifier_to_likelihood-ratio}
    r(\thetab, \Zvec) = \frac{c^*(\thetab, \Zvec)}{1 - c^*(\thetab, \Zvec)}, \quad \thetab \in \Theta,~ \Zvec \in \mathcal{Z}.
\end{equation}
In the typical case that $c^*(\cdot, \cdot)$ is unavailable, it can be approximated by adopting a flexible parametric representation for $c(\cdot, \cdot)$ and maximizing a Monte Carlo approximation of the objective function in Equation~\ref{eqn:optimal_classifier}.  Specifically, let $c_{\gammab}(\cdot, \cdot)$ denote a binary classifier parameterized by $\gammab$, typically a neural network with a sigmoid output activation function, although other representations are possible \citep[e.g., a logistic regression based on user-specified or learned summary statistics;][]{Dinev_2018, Thomas_2022_ratio_estimation}. Then, the Bayes classifier may be approximated by $c_{\gammab^*}(\cdot, \cdot)$, where
\begin{equation}\label{eqn:optimal_classifier_approx}
 \gammab^* = \argmax_\gammab \sum_{i=1}^N \Big[\log\{c_{\gammab}(\thetab^{(i)}, \Zvec^{(i)})\} +  \log\{1 - c_{\gammab}(\thetab^{(i)}, \Zvec^{\{\pi(i)\}})\} \Big],
\end{equation}
where $\thetab^{(i)} \sim p(\thetab)$, $\Zvec^{(i)} \sim p(\Zvec \mid \thetab^{(i)})$ and $\pi(\cdot)$ is a random permutation of $\{1, \dots, N\}$. Figure~\ref{fig:Bayes_classifier} demonstrates this learning task on the simple model $Z \mid \theta \sim \textrm{Gau}(\theta, \theta^2)$, $\theta \sim \textrm{Unif}(0, 1)$, for $c_{\gammab}(\cdot, \cdot)$ a fully-connected neural network with six layers, each of width 64.  

Once trained, the classifier $c_{\gammab^*}(\cdot, \cdot)$ may be used to quickly approximate the likelihood-to-evidence ratio via Equation~\ref{eqn:optimal_classifier_to_likelihood-ratio}, as summarized in the graphical representation shown in Figure~\ref{fig:NeuralClassifier}. Inference based on an amortized approximate likelihood-to-evidence ratio may proceed as discussed in the introduction to Section~\ref{sec:likelihood_and_ratio}, namely, in a frequentist setting via maximum likelihood estimation and likelihood ratios \citep[e.g.,][]{Walchessen_2023_neural_likelihood_surfaces}, and in a Bayesian setting by facilitating the computation of transition probabilities in Hamiltonian Monte Carlo and MCMC algorithms \citep[e.g.,][]{Hermans_2020, Begy_2021}. Furthermore, an approximate posterior distribution can be obtained via the identity ${p(\thetab \mid \Zvec)} = p(\thetab) r(\thetab, \Zvec)$, and sampled from using standard sampling techniques (e.g., \citeauthor{Thomas_2022_ratio_estimation} \citeyear{Thomas_2022_ratio_estimation}; see also Section~\ref{sec:illustration}). 
Finally, if one also has an amortized likelihood approximator (Section~\ref{sec:NLE}), then one may approximate the model evidence via $p(\Zvec) = p(\Zvec \mid \thetab)/r(\thetab, \Zvec)$.

\begin{figure}[t!]
\begin{center}
  \begin{tikzpicture}[>=latex]
    
    \node[circle, draw, minimum size=1.1cm, inner sep=0pt] (Z) at (0,0) {$\thetab, \Zvec$};
    \node[rectangle, draw, minimum size=1cm, inner sep=0pt, fill=black, text=white] (NN) at (3,0) {\textbf{NN}};
    \node[circle, draw, minimum size=1.1cm, inner sep=0pt] (c) at (6,0) {$c$};
    \node[circle, draw, minimum size=1.1cm, inner sep=0pt] (r) at (9,0) {$r$};
    
    \draw[->] (Z) to (NN);
    \draw[->] (NN) to (c);
    \draw[->] (c) to (r);
    
    \draw[decorate,decoration={brace,amplitude=10pt,mirror}] 
([shift={(-0.2,-0.3)}]Z.south west) -- ([shift={(0.2,-0.3)}]Z.south east) 
node[midway,below=10pt,align=center] {Parameters and\\ Data};
    \draw[decorate,decoration={brace,amplitude=10pt,mirror}] 
([shift={(-0.2,-0.2)}]NN.south west) -- ([shift={(0.2,-0.2)}]NN.south east) 
node[midway,below=10pt,align=center] {Neural \\ Classifier};
    \draw[decorate,decoration={brace,amplitude=10pt,mirror}] 
([shift={(-0.2,-0.3)}]c.south west) -- ([shift={(0.2,-0.3)}]c.south east) 
node[midway,below=10pt,align=center] {Class \\ Probability};
    \draw[decorate,decoration={brace,amplitude=10pt,mirror}] 
([shift={(-0.2,-0.3)}]r.south west) -- ([shift={(0.2,-0.3)}]r.south east) 
node[midway,below=10pt,align=center] {Likelihood-to-Evidence \\ Ratio};
    \end{tikzpicture}
    \end{center}
    \caption{Graphical representation of a neural classifier that takes both data and parameters as input and that outputs a class probability that is then used to approximate the likelihood-to-evidence ratio.\label{fig:NeuralClassifier}}
  \end{figure}

\subsubsection{Variants of the target ratio}\label{sec:lik-ratio-variants}

Several variants of the learning task described above have been investigated, with a variety of different aims. One is that of safeguarding against over-optimistic inferences when the true Bayes classifier $c^*(\cdot,\cdot)$ cannot be well approximated by the neural classifier $c_{\gammab}(\cdot,\cdot)$; in this situation it is widely acknowledged that one should err on the side of caution and make $c_{\gammab}(\cdot,\cdot)$ conservative 
\citep{Hermans_2022}. \cite{Delaunoy_2022_balanced_NRE} propose a way to do this by encouraging $c_{\gammab}(\cdot,\cdot)$ to satisfy the balance condition
\begin{equation}\label{eqn:NRE:balance}
    \E_{(\thetab', \Zvec')'} \{c_{\gammab}(\thetab, \Zvec)\} + \E_{(\tilde{\thetab}',\tilde{\Zvec}')'} \{c_{\gammab}(\tilde{\thetab}, \tilde{\Zvec})\} = 
    1;
\end{equation}
it is straightforward to see that the Bayes classifier under equal class probability, $c^*(\cdot,\cdot)$, satisfies the condition. One can show that any classifier satisfying Equation~\ref{eqn:NRE:balance} also satisfies
$$\E_{(\thetab', \Zvec')'} \left\{\frac{c^*(\thetab, \Zvec)}{c_{\gammab}(\thetab, \Zvec)}\right\} \geq 1, \textrm{~~~~and~~~~} \E_{(\tilde{\thetab}',\tilde{\Zvec}')'} \left\{\frac{1 - c^*(\tilde{\thetab}, \tilde{\Zvec})}{1 - c_{\gammab}(\tilde{\thetab}, \tilde{\Zvec})}\right\} \geq 1,$$
which informally means that the classifier tends to produce class-probability estimates that are smaller (i.e., less confident and more conservative) than those produced by the (optimal) Bayes classifier. To encourage the trained estimator to satisfy the balancing condition, and hence be conservative, one may append to the objective function in Equation~\ref{eqn:optimal_classifier} the penalty term
 \begin{equation}\label{eqn:NRE:balance_penalty}
    \lambda \Big[\E_{(\thetab', \Zvec')'} \{c(\thetab, \Zvec)\} + \E_{(\tilde{\thetab}',\tilde{\Zvec}')'} \{c(\tilde{\thetab}, \tilde{\Zvec})\} - 
    1\Big]^2, \quad \lambda > 0,
\end{equation}
which equals zero when the balancing condition is satisfied. Since the Bayes classifier with equal class probabilities is balanced, this penalty is only relevant when the Bayes classifier cannot be approximated well using the neural network.

Another research focus is improving computational efficiency for amortized ratio approximators. For example, approximating the likelihood ratio ${\tilde{r}(\thetab_0, \thetab_1, \Zvec) \equiv p(\Zvec \mid \thetab_0)/{p(\Zvec \mid \thetab_1)}}$  $ =  r(\thetab_0, \Zvec)/r(\thetab_1, \Zvec)$ 
requires two forward passes through a neural classifier $c_{\gammab^*}(\cdot, \cdot)$. Recently, however, \citet{Cobb_2023_directed_amortized_likelihood_ratio_estimation} proposed to construct an amortized approximation of $\tilde{r}(\cdot, \cdot, \cdot)$ using a single neural network that takes two parameter vectors as input. This only requires one forward pass, improving computational efficiency, though this benefit may be offset if a larger neural network is needed to learn the more complicated mapping. Finally, several approaches have been developed to construct amortized ratio approximators for a subset of $\thetab$. These marginal methods include introducing a binary mask as input to encode the desired subset of parameters \citep{rozet2021arbitrary}, and constructing separate approximators for each subset of interest \citep{ Miller_2021_truncated_marginal_NRE, swyft}.

\section{SOFTWARE AND EXAMPLE}\label{sec:software}

Although amortized neural inference is a relatively new field, many easy-to-use software packages are publicly available. We outline these in Section~\ref{sec:software_only} and demonstrate their application through a simple example in Section~\ref{sec:illustration} where the model parameters are easily inferred using MCMC. An additional example with a model unsuitable for MCMC is given in Section~\ref{sec:inv-max-stable} of the Supplemental Appendix. Code for these two examples is available at \url{https://github.com/andrewzm/Amortized_Neural_Inference_Review/}.

\subsection{Software}\label{sec:software_only}

Most publicly available software is written in \proglang{Python}, and leverages the deep-learning libraries \pkg{TensorFlow} \citep{tensorflow} and \pkg{PyTorch} \citep{pytorch}. Software is also available in the numerical programming language \proglang{Julia} and its deep-learning library \pkg{Flux} \citep{flux}. There are \proglang{R} interfaces to several of these libraries written in other languages. Software evolves quickly; the following description is based on availability and functionality as of early 2024.

The package \pkg{NeuralEstimators} \citep{Sainsbury_2024} helps construct neural Bayes estimators (Section~\ref{sec:neural_point_estimation}) using arbitrary loss functions, and likelihood-to-evidence ratio approximators (Section~\ref{sec:NRE}). The package is written in \proglang{Julia}, leverages the deep-learning library \pkg{Flux}, and is accompanied by a user-friendly \proglang{R} interface. The package is designed to deal with exchangeable data, such as independent realizations from a spatial process model. The package also implements methods for bootstrap-based uncertainty quantification and for handling censored \citep{Richards_2023} and missing \citep{Wang_2022_neural_missing_data} data.

\begin{marginnote}
   \entry{Censored data}{Data whose precise values are not known, but which are known to lie in some interval. Multivariate censored data often lead to intractable likelihood functions.}
   \entry{Missing data}{Data that have not been observed, and that one often must predict or impute.}
 \end{marginnote}

The package \pkg{sbi} \citep{tejero-cantero2020sbi}, short for simulation-based inference, is a \proglang{Python} package built on \pkg{PyTorch} that provides methods for targeting the posterior distribution (Section~\ref{sec:forwardKL}), the likelihood function (Section~\ref{sec:NLE}), or the likelihood-to-evidence ratio (Section~\ref{sec:NRE}), with posterior inference using the likelihood or likelihood-to-evidence ratio facilitated with MCMC sampling, rejection sampling, or (non-amortized) variational inference. It implements both amortized and sequential methods, the latter aimed at improving simulation efficiency.

The \pkg{PyTorch}-based package \pkg{LAMPE} \citep{lampe}, short for likelihood-free amortized posterior estimation, focuses on amortized methods for approximating the posterior distribution (Section~\ref{sec:forwardKL}) or the likelihood-to-evidence ratio (Section~\ref{sec:NRE}), with posterior inference using the likelihood-to-evidence ratio facilitated with MCMC or nested sampling. The package allows training data to be stored on disk and dynamically loaded on demand, instead of being cached in memory; this technique facilitates the use of very large datasets.

The \pkg{TensorFlow}-based package \pkg{BayesFlow} \citep{Radev_2023_Bayesflow_package} implements methods for approximating the posterior distribution (Section~\ref{sec:forwardKL}) and the likelihood function (Section~\ref{sec:NLE}), possibly  jointly \citep{Radev_2023}; detecting model misspecification \citep{Schmitt_2021,Schmitt_2024}; and performing amortized model comparisons via posterior model probabilities or Bayes factors. Recently, neural Bayes estimators (Section~\ref{sec:neural_point_estimation}) have also been incorporated into the package. The package is well documented and provides a user-friendly application programming interface. 

The \pkg{PyTorch}-based package \pkg{swyft} \citep{swyft} implements methods for estimating likelihood-to-evidence ratios for subsets of the parameter vector, using amortized and sequential training algorithms. To facilitate the use of datasets that are too large to fit in memory, the package also allows data to be stored on disk and dynamically loaded.

The above software packages provide the tools necessary to easily train the required neural networks from scratch, but are general-purpose. Future amortized inference tools may be bundled with packages primarily designed around a specific modeling framework. For instance, a package that implements a particular model could include a pre-trained neural network, or several such neural networks, designed to make fast inferences for that model. This could be a natural evolution since there are incentives for the developers of such packages to make inference straightforward for increased accessibility and uptake by end users.

\begin{marginnote}
  \entry{Pre-trained neural network}{A neural network trained for a generic task, which is used as a starting network when training for specialized tasks (e.g., in our context, for a slightly different model or prior distribution).}
\end{marginnote}

\subsection{Example}\label{sec:illustration}

Protocols for benchmarking amortized inference methods are still in their infancy \citep{Lueckmann_2021}. To showcase a few techniques discussed in this review, here we consider a model with only one unknown parameter, where the neural networks are straightforward to train without the need for high-end GPUs, and  for which MCMC is straightforward to implement for comparison. We consider a spatial Gaussian process with exponential covariance function with unit variance and unknown length scale $\theta > 0$. We assume that data $\Zvec$ are on a ${16 \times 16}$ gridding, $\mathcal{D}^G$, of the unit square $\mathcal{D}=[0,1]^2$. The process model is therefore given by ${\Zvec \mid \theta \sim \textrm{Gau}\{\zerob, \Sigmamat(\theta)\}}$, where ${\Sigmamat \equiv (\exp(-\|\svec_i - \svec_j\|/\theta): \svec_i, \svec_j \in \mathcal{D}^G)}$, and we let ${\theta \sim \textrm{Unif}(0, 0.6)}$. We train the neural networks on 160,000 draws from $p(\theta)$ and $p(\Zvec\mid\theta)$, and test all methods on 1,000 independent draws.

The techniques we consider, their acronyms, and their software implementations are outlined in Table~\ref{tab:software} in the Supplemental Appendix. The gold standard is provided by a (non-amortized) Metropolis--Hastings algorithm, \texttt{MCMC}, run on the 1,000 test datasets. We implement a neural Bayes estimator, \texttt{NBE} (Section~\ref{sec:neural_point_estimation}), using \pkg{NeuralEstimators}, targeting the posterior mean, the posterior 5th percentile and the posterior 95th percentile. We implement amortized posterior inference via forward KL minimization, \texttt{fKL} (Section~\ref{sec:forwardKL}), using \pkg{BayesFlow} with a normalizing flow model for the posterior distribution constructed using affine coupling blocks. In order to ensure that the approximate posterior densities are zero outside the prior support of  $[0, 0.6]$, we  make inference on $\tilde\theta \equiv \Phi^{-1}(\theta/0.6) \sim \textrm{Gau}(0,1)$, and obtain samples from the posterior distribution of $\theta$ through the inverse $\theta = 0.6\Phi(\tilde\theta)$, where $\Phi(\cdot)$ is the standard normal cumulative distribution function.
\begin{marginnote}
\entry{Affine coupling blocks}{An invertible transformation, where the input is partitioned into two blocks. One of the blocks undergoes an affine transformation that depends on the other block.}
\end{marginnote}

We implement three types of reverse-KL methods using \pkg{TensorFlow}: \texttt{rKL1}, \texttt{rKL2}, and \texttt{rKL3}, which differ in the likelihood function they use. All three target an approximate Gaussian posterior distribution of $\tilde{\theta} \equiv \log (\theta / (0.6 - \theta))$; samples from the approximate (non-Gaussian) posterior distribution of $\theta$ are obtained by back-transforming samples drawn from the approximate Gaussian posterior distribution of $\tilde{\theta}$. The first variant, \texttt{rKL1}, uses the true likelihood function (Section~\ref{sec:NVI}); the second, \texttt{rKL2}, uses a synthetic one constructed with an ``expert'' summary statistic, in this case given by the mean of the squared differences between neighboring pixels in $\Zvec$ (Section~\ref{sec:NVI-synth}); and the third, \texttt{rKL3}, uses a synthetic one constructed using a summary statistic found by maximizing mutual information (Section~\ref{sec:explicit_summ}). We also consider the neural ratio estimation method, \texttt{NRE}, implemented in the package \pkg{sbi} (Section~\ref{sec:NRE}); we use the amortized ratio to quickly evaluate the posterior distribution on a fine gridding of the parameter space, from which we then draw samples. For all approaches we use similar architectures, largely based on a two-layer CNN. Each neural network took a few minutes to train using the CPU of a standard laptop.

\begin{figure}

  \begin{minipage}{.35\textwidth}
   \subfloat{\includegraphics[width=\textwidth]{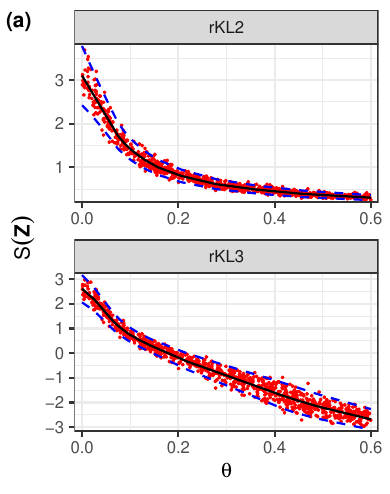}}
\end{minipage}
\hfill
 \begin{minipage}{.65\textwidth}
    \subfloat{\includegraphics[width=\textwidth]{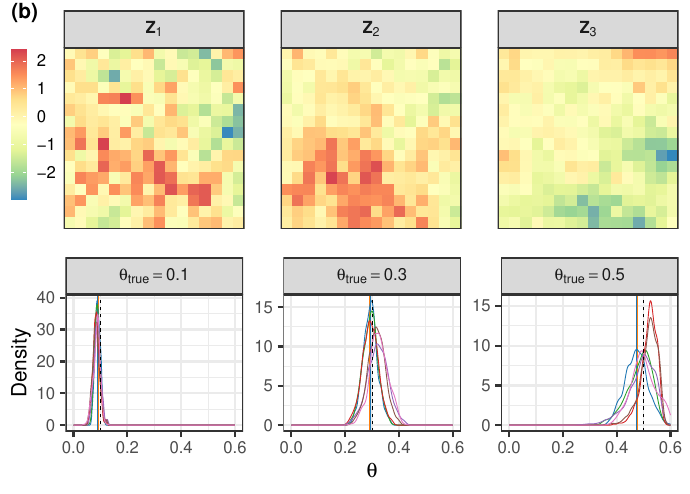}\label{fig:sub_b}}
\end{minipage}
\hfill\\

\vspace{-0.2in}

\begin{minipage}{\textwidth}
    \subfloat{\includegraphics[width=\textwidth]{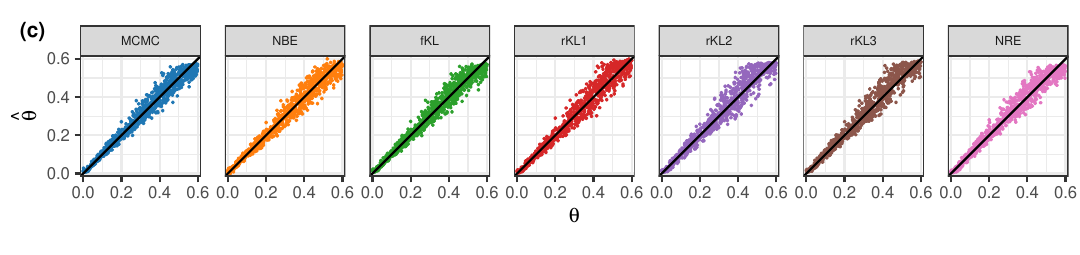}\label{fig:sub_a}}
  \end{minipage}\hfill    \\

\vspace{0.2in}

\caption{Results from the illustration of Section~\ref{sec:illustration}. Panel (a) plots the summary statistics of the test data (red), together with the mean (solid black line) and 2$\sigma$ bands constructed from the fitted neural binding functions. The top sub-panel shows the expert summary statistic (used in \texttt{rKL2}), while the bottom sub-panel shows the statistic constructed by maximizing mutual information (used in \texttt{rKL3}). Panel (b) shows the results from all methods on three test datasets. The top sub-panels show the data, while the bottom sub-panels show estimates (solid vertical lines, \texttt{MCMC} posterior mean and \texttt{NBE}, which largely overlap) and the inferred posterior distributions (solid lines, all methods except \texttt{NBE}), as well as the true parameter value (dashed line). Panel (c) plots parameter point estimates (in this case the posterior mean) against the true parameter values in the test datasets for all methods. The colours of the lines in panel (b) correspond to the colours used for the different methods in panel (c). 
\label{fig:results}}
  \end{figure}

We summarize results in Figure~\ref{fig:results} and Table~\ref{tab:results}, where we assess the methods on the test data using scoring rules \citep{Gneiting_2007} and 90\% empirical coverages \citep[e.g.,][]{Hermans_2022}. Figure~\ref{fig:results}a shows the summary statistics for \texttt{rKL2} and \texttt{rKL3}, along with the inferred mean $\mu_{\taub^*}(\theta)$ and the $2\sigma_{\taub^*}(\theta)$ interval used to construct the synthetic likelihood. The expert summary statistic is non-linear and, for large $\theta$, a broad range of parameter values lead to a small expert summary statistic. The statistic constructed by maximizing the mutual information is largely linear over the entire support of $p(\theta)$; as seen from Table~\ref{tab:results} this leads to slightly better performance. Figure~\ref{fig:results}b shows the results from applying the methods to three (test) spatial fields, $\Zvec_1, \Zvec_2$, and $\Zvec_3$. All methods perform well, with the posterior variance increasing with $\theta$ as expected. Differences between the approaches are mostly evident for larger $\theta$, where inferences are more uncertain.
Figure~\ref{fig:results}c plots the posterior means against the true values; all methods again perform as expected, with lower variance for small $\theta$ and large variance for large $\theta$, where the length scale is large in comparison to the size of the spatial domain. 

\begin{table}[t!]
   \caption{Evaluation of the methods in Section~\ref{sec:illustration} on test data using the root median squared prediction error (RMSPE), the median 90\% interval score (MIS90), the median continuous-ranked probability score (MCRPS), and the 90\% empirical coverage (COV90). We use the median instead of the mean since the distribution of the metrics is highly skewed; see Figure~\ref{fig:score_distribution} in the Supplemental Appendix. All scores excluding COV90 (which should be close to 90\%) are negatively oriented (lower is better). MCRPS is not available with our implementation of the NBE.\label{tab:results}}

   \vspace{0.2in}

   \centering
   \begin{tabular}{rrrrrrrr} \hline
 & \texttt{MCMC} & \texttt{NBE} & \texttt{fKL} & \texttt{rKL1} & \texttt{rKL2} & \texttt{rKL3} & \texttt{NRE} \\ 
  \hline
RMSPE $(\times 10^2)$ & 1.50 & 1.56 & 1.50 & 1.64 & 1.88 & 1.78 & 1.70 \\ 
  MIS90 $(\times 10^2)$ & 9.55 & 10.62 & 9.89 & 9.80 & 11.36 & 9.07 & 10.99 \\ 
  MCRPS $(\times 10^2)$ & 1.41 & NA  & 1.46 & 1.36 & 1.61 & 1.37 & 1.59 \\ 
  COV90 $(\%)$ & 89.60 & 87.70 & 90.60 & 77.70 & 87.10 & 81.60 & 91.30 \\ 
   \hline

   \end{tabular}
\end{table}

As seen from Table~\ref{tab:results}, all methods perform slightly worse overall than 
 \texttt{MCMC} due to the amortization gap (\texttt{NBE}, \texttt{fKL}, \texttt{NRE}), the use of an inflexible approximation to the posterior distribution (the \texttt{rKL} variants), or both. However, the significance of amortization lies in the speed with which these inferences can be obtained, and the general class of models the neural methods apply to. In this example, one run of \texttt{MCMC} required around one minute of compute time to generate 24,000 samples (which we thinned down to 1,000), while the neural methods only required a few dozen milliseconds each to yield approximate posterior inferences. 

\begin{marginnote}
  \entry{Continuous ranked probability score}{Given a forecast distribution $F$ and an observation $y$, the continuous ranked probability score is $\textrm{CRPS}(F,y) \equiv \int_{-\infty}^\infty \left\{ F(x) \right. - $\\ $~~~~~~\left.\mathbbm{1}(y \le x)\right\}^2 \intd x$.}
  \entry{Interval score}{Given the lower and upper quantiles, $y_l$ and $y_u$, of a forecast distribution, and an observation $y$, the $(1 - \alpha)\times 100\%$ interval score is $\textrm{IS}_\alpha(y_l,y_u;y) \equiv (y_u - y_l) + \\ {\frac{2}{\alpha}(y_l - y)\mathbbm{1}(y < y_l)} + \frac{2}{\alpha}(y - y_u)\mathbbm{1}(y > y_u)$.} 
\end{marginnote}

\section{EPILOGUE}\label{sec:epilogue}
Amortized neural inference methods are new, but they have already shown enormous potential for accurate and fast statistical inference in a wide range of applications with models that either are defined explicitly but are computationally intractable, 
 or that are defined implicitly through a stochastic generator. 
The amortized nature of the methods presented in this review article, enabled by neural networks for quick evaluation, opens a new era in statistical inference, in which one can construct and share pre-trained inference tools for off-the-shelf use with new data at negligible cost. The field is rapidly expanding and, to preserve the review's focus, many relevant topics and background materials were not discussed in depth. For further reading on related subjects, see Section~\ref{sec:further_reading_supp} of the Supplemental Appendix.

Several challenges remain for future research. From a theoretical viewpoint, we need a better understanding of the asymptotic properties of neural inference tools (e.g., consistency, rate of convergence based on network architecture and training set size) to guide their design and establish rigorous implementation strategies. From a methodological viewpoint, several avenues remain unexplored. For example, recovering distributions of random effects in hierarchical models could be achieved by combining amortized methods for latent variable inference \citep[e.g.,][]{Liu_2019} with the parameter inference methods discussed in this review. Additionally, some methods like ABC can also be amortized \citep[e.g.,][]{Mestdagh_2019}, and it is not yet clear what advantages these have, if any, over the approaches discussed in this review. More traditional branches of statistics that can benefit from these advances include analyses of survey data, crop yield, and experimental design.  Amortized neural inference approaches also enable online frequentist or Bayesian inference where data arrive sequentially, and where parameters need to be tracked. We anticipate that several fields in statistics will soon be impacted by this relatively new enabling technology.

\section*{DISCLOSURE STATEMENT}
The authors are not aware of any affiliations, memberships, funding, or financial holdings that might be perceived as affecting the objectivity of this review.

\section*{ACKNOWLEDGEMENTS}

All authors were supported by the King Abdullah University of Science and Technology (KAUST) Opportunity Fund Program ORFS-2023-OFP-5550.2. This material is based upon work supported by the Air Force Office of Scientific Research under award number FA2386-23-1-4100 (A.Z.-M.). M.S.-D.'s research was additionally supported by an Australian Government Research Training Program Scholarship, a 2022 Statistical Society of Australia (SSA) top-up scholarship, the KAUST Office of Sponsored Research (OSR) under Award No.~OSR-CRG2020-4394, and R.H.'s baseline funds.

\bibliographystyle{paper} 
\bibliography{paper}

\end{document}